\documentclass[letterpaper]{article} 
\usepackage{aaai2026}  
\usepackage{times}  
\usepackage{helvet}  
\usepackage{courier}  
\usepackage[hyphens]{url}  
\usepackage{graphicx} 
\usepackage{natbib}  
\usepackage{caption} 
\usepackage{algorithm}
\usepackage{algorithmic}

\usepackage{amsmath,amssymb}
\usepackage{booktabs}
\usepackage{xcolor}
\usepackage{multirow}
\usepackage{tikz} 
\usepackage{xcolor} 

\usepackage{newfloat}
\usepackage{listings}
\DeclareCaptionStyle{ruled}{labelfont=normalfont,labelsep=colon,strut=off} 
\floatstyle{ruled}
\newfloat{listing}{tb}{lst}{}
\floatname{listing}{Listing}
\title{XLinear: A Lightweight and Accurate MLP-Based Model for Long-Term Time Series Forecasting with Exogenous Inputs}
\author{
    Xinyang Chen\textsuperscript{\rm 1}\equalcontrib,
    Huidong Jin\textsuperscript{\rm 2}\equalcontrib,
    Yu Huang\textsuperscript{\rm 1,3,4}\thanks{Corresponding authors.},
    Zaiwen Feng\textsuperscript{\rm 1,3,4}$^\dagger$
}
\affiliations{
    \textsuperscript{\rm 1}College of Informatics, Huazhong Agricultural University, Wuhan, Hubei, China \\
    \textsuperscript{\rm 2}
Statistical Machine Learning, 
Data61, CSIRO, Canberra Australia\\
    \textsuperscript{\rm 3}Hubei Key Laboratory of Agricultural Bioinformatics, Wuhan, Hubei, China\\
    \textsuperscript{\rm 4}Engineering Research Center of Agricultural Intelligent Technology, Ministry of Education, China\\
    Warren.Jin@csiro.au, \{Yhuang, Zaiwen.Feng\}@mail.hzau.edu.cn
}

\begin{document}

\maketitle

\begin{abstract}
Despite the prevalent assumption of uniform variable importance in long-term time series forecasting models, real-world applications often exhibit asymmetric causal relationships and varying data acquisition costs. Specifically, cost‐effective exogenous data (e.g., local weather) can unilaterally influence dynamics of endogenous variables, such as lake surface temperature. Exploiting these links enables more effective forecasts when exogenous inputs are readily available. Transformer-based models capture long-range dependencies but incur high computation and suffer from permutation invariance. Patch-based variants improve efficiency yet can miss local temporal patterns. To efficiently exploit informative signals across both the temporal dimension and relevant exogenous variables, this study proposes XLinear, a lightweight time series forecasting model built upon Multi-Layer Perceptrons (MLPs). XLinear uses a global token derived from an endogenous variable as a pivotal hub for interacting with exogenous variables, and employs MLPs with sigmoid activation to extract both temporal patterns and variate-wise dependencies. Its prediction head then integrates these signals to forecast the endogenous series. We evaluate XLinear on seven standard benchmarks and five real-world datasets with exogenous inputs. Compared with state-of-the-art models, XLinear delivers superior accuracy and efficiency for both multivariate forecasts and univariate forecasts influenced by exogenous inputs.
\end{abstract}
\begin{links}
    \link{Code}{https://github.com/Zaiwen/XLinear.git}
\end{links}

\section{Introduction}

Time series forecasting has shown strong value across domains 
such as
financial analysis~\cite{liu2023financial}, irrigation scheduling~\citep{shao2019new,shao2025comprehensive}, population flows~\citep{bakar2018spatio}, yield forecasting~\citep{jin2022improving}, and environmental science~\citep{genova2025advancing,bakar2015spatiodynamic,bakar2016hierarchical,kokic2013improved,jin2011towards}. The predictability of time series data lies in two main aspects: persistent temporal patterns and cross-variate information~\cite{chen2023tsmixer}. In practice, actual forecasts require modeling both the endogenous time series and relevant exogenous inputs, which often influence the target dynamics rather than the other way around. For instance, crop 
growth for a given genotype is influenced by environmental 
and management factors over time~\citep{jin2022improving}, whereas lake surface or crop canopy temperature is primarily driven by local weather~\citep{shao2019new,shao2025comprehensive}.
\begin{figure}[t]
    \begin{centering}	
    \includegraphics[width=0.45\textwidth]{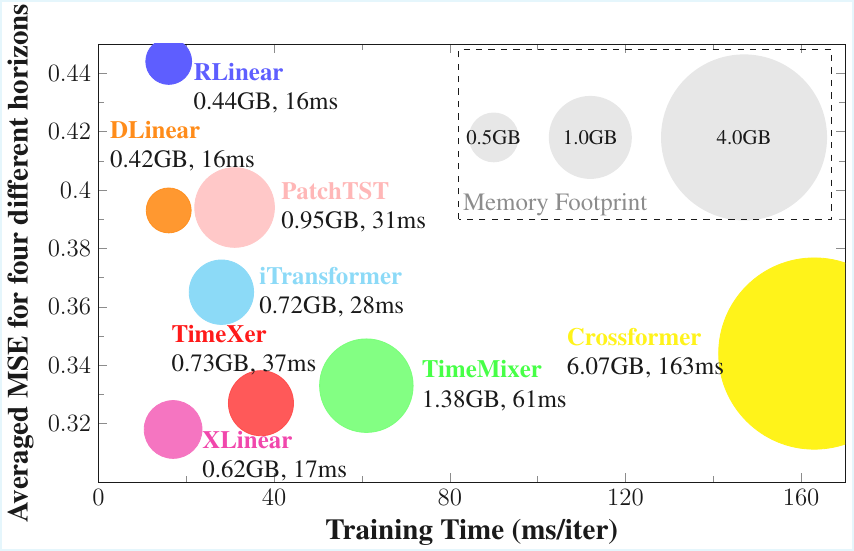}
    \caption{Model performance on the Electricity dataset for univariate forecasting with 320 exogenous variables. TimesNet and TiDE are excluded for clarity due to slow training.}
        \label{fig:effience}
    \end{centering}
\end{figure}
With the advent of deep learning, the field has witnessed transformative advancements, particularly with Transformer~\cite{vaswani2017attention} emerging as the prevailing modeling paradigm. Leveraging its multi-head self-attention mechanism, Transformer excels at capturing long-range dependencies in sequential data, establishing itself as a benchmark for long-term time series forecasting. Early Transformer-based models~\cite{li2019enhancing,zhou2021informer,wu2021autoformer,zhou2022fedformer} focused on optimizing the self-attention mechanism while paying little attention to cross-variable dependencies. By assuming channel-independence, PatchTST~\cite{nie2022time} leveraged patch-based attention to gain high efficiency. Crossformer~\cite{zhang2023crossformer} and iTransformer~\cite{liu2023itransformer} focused on cross-channel modeling but both have limitations in temporal modeling: Crossformer's unaligned patch interactions may introduce noise, while iTransformer weakly captures temporal dependencies. The recent TimeXer~\cite{wang2024timexer} model established a new state-of-the-art (SOTA) benchmark for both multivariate and univariate time series forecasting with exogenous drivers, by effectively extracting information from endogenous time series via self-attention and from exogenous drives via cross-attention. However, patch-based Transformer models excessively rely on the patching mechanism to achieve desirable performance, which limits their applicability in forecasting tasks unsuitable for patching~\cite{luo2024deformabletst}. Additionally, their permutation-invariant self-attention mechanism suffers from temporal information loss~\cite{tang2025unlocking}. Inspired by its global tokens for endogenous variables, we will propose a more efficient and effective model in this study.

Since the emergence of lightweight models for long-term time series forecasting~\cite{zeng2023transformers}, doubts have been raised about the actual effectiveness of Transformer, which also inspired our research and model design. Among them, TSMixer~\cite{chen2023tsmixer} models both temporal and cross-variable dependencies simultaneously, but its forecast accuracy does not show significant improvement over PatchTST, indicating its cross-variable modeling limitation. DLinear~\cite{zeng2023transformers}, TimeMixer~\cite{wang2024timemixer}, and xPatch~\cite{stitsyuk2025xpatch} decompose the seasonal-trend components in the temporal dimension of the sequence and capture complex temporal patterns through specific processing methods. Although these models can capture temporal dependencies, for real-world forecast applications, crucial information from exogenous drives cannot be ignored. 

Pre-trained foundation time series models have been made impressive progress in the last few years. Almost all of them are Transformer-based~\cite{kottapalli2025foundation}. Their time series forecast accuracy remains limited on new and unseen data sets or some in-domain data sets~\cite{shi2025time}, comparing with full-shot SOTA models like TimeXer and TimeMixer. There are various other deep models in the literature, such as those based on convolution mechanisms~\cite{wu2022timesnet}, and graph neural networks (GNN)~\cite{yi2023fouriergnn,cai2024msgnet}. Their forecast accuracy is not as good as TimeXer~\cite{wang2024timexer} and our proposed model.

Inspired by the advances of TimeXer and lightweight models, this study proposes XLinear, a lightweight model that integrates a Multi-Layer Perceptron (MLP) with a sigmoid activation function as a gating module.
Unlike TimeXer, which employs patch-level self-attention and variate-level cross-attention to capture temporal and inter-variable dependencies, XLinear utilizes unified gating modules that sequentially filter features along the temporal and variable dimensions, achieving more precise and efficient dependency modeling.
The prediction head employs a cross-channel fusion mechanism to synergistically integrate bidimensional information, effectively mitigating noise arising from interactions between features of different dimensions. Although XLinear comprises only two sets of MLP-based feature extractors, it strikes a strong balance between accuracy and efficiency. As shown in Fig. \ref{fig:effience}, in the benchmark test on the Electricity dataset (with a unified batch size of 4), XLinear matches the training speed of the simple yet effective DLinear and RLinear, consumes less memory than SOTA models, and delivers the highest forecast accuracy among all 10 comparative models.

The core contributions of this study are as follows: 
\begin{description}
    \item[Bridging the Efficiency-Accuracy Gap:] We address the critical challenge in time series forecasting, especially with exogenous inputs, of balancing computational efficiency (a strength of MLP-based models) with high forecasting accuracy (a hallmark of patch-based Transformers). Our work directly tackles this by explicitly incorporating exogenous inputs and effectively reconciling complex temporal and cross-variable dependencies.
    \item[Introducing XLinear, a Novel MLP-Based Model:] Its core innovation lies in a novel gating module that uses an MLP with a sigmoid activation function that enables XLinear to efficiently boost predictive accuracy by fully leveraging informative exogenous signals through cross-variable dependencies, facilitated by global tokens derived from endogenous sequences.
    \item[Demonstrating Superior Performance:] Through extensive tests on 12 diverse datasets, XLinear consistently outperforms SOTA models in both accuracy and efficiency. Notably, XLinear achieves at least 30\% faster training speeds than these efficient Transformers. 
\end{description}

\begin{figure*}[t]
    \centering
    \includegraphics[width=0.99\textwidth]{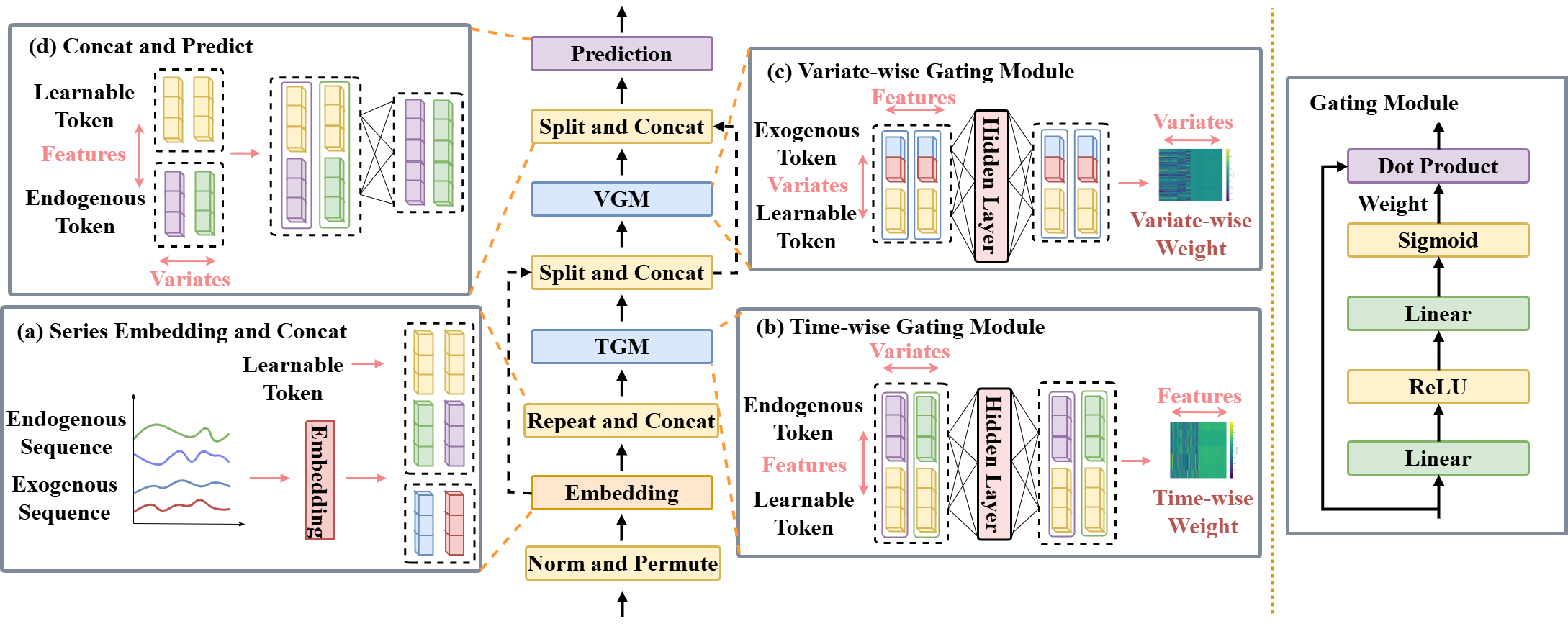}
    \caption{Architecture of XLinear. The left panel illustrates the core processing pipeline: (a) Synchronous processing of endogenous and exogenous sequences via a joint embedding layer, with learnable global representation tokens introduced for endogenous variables; (b) Time-wise Gating Module (TGM) enhances key temporal features and suppresses noise in the endogenous sequences while transferring critical temporal characteristics to the global tokens; (c) Variate-wise Gating Module (VGM) leverages the global tokens of the endogenous sequences to interact with the exogenous sequences across dimensions, extracting implicit cross-variable dependencies; (d) Cross-variable information carried by each global token is concatenated with its associated temporally enhanced endogenous sequence, followed by generating time-series predictions for endogenous variables via the prediction head. The right gating module illustrates the implementation method of TGM and VGM.}
    \label{fig:framework}
\end{figure*}

\section{Method}
Previously, MLP-based models have demonstrated strong capabilities in capturing internal time series dependencies. Building on this foundation, we design an MLP-based gating module to efficiently capture dependencies across both temporal and variable dimensions, enabling selective feature filtering. To effectively integrate crucial information across these dimensions, inspired by TimeXer~\cite{wang2024timexer}, we introduce learnable global tokens for each endogenous variable, serving as hubs for interactions with exogenous variables and thereby facilitating causal information transfer. The separation of global and temporal tokens effectively mitigates cross-variable noise, while their integration enhances feature representation and forecasting performance. The architecture of XLinear is illustrated in Fig. \ref{fig:framework}.

\subsection{Forward Process}
Given the historical observation sequence of the endogenous variables \( X_{1:T} = \{X_{1:T}^{(1)}, X_{1:T}^{(2)}, \ldots, X_{1:T}^{(M)}\} \), where \( X_{1:T}^{(j)} \) denotes the sequence of the \( j\)-th endogenous variable, and the sequence of exogenous variables up to time \( T \), denoted as \( E_{1:T} = \{E_{1:T}^{(1)}, E_{1:T}^{(2)}, \ldots, E_{1:T}^{(C)}\} \), where \( E_{1:T}^{(i)} \) represents the sequence of the \( i\)-th exogenous variable, the model is designed to predict the values of the endogenous variable sequence for the subsequent \( S \) time steps.
The entire process can be described as:
\begin{equation}
\hat{X}_{(T+1):(T+S)} = \boldsymbol{\mathrm{F}}(X_{1:T}, E_{1:T}) \in \mathbb{R}^{M \times S}.
\end{equation}
\( \boldsymbol{\mathrm{F}(\cdot)} \) is a mapping learned by the model during training, enabling the conversion of input sequences to \( S \)-step-ahead predictions.

\subsection{Embedding}
To capture latent features in time-series data, we jointly embed the RevIN-processed \cite{kim2021reversible} endogenous sequence $X_{1:T}$ and exogenous sequence $E_{1:T}$. The process is:
\begin{equation}
X_{\text{endo}}, E_{\text{exo}} = \text{Embedding}(X_{\text{1:T}}, E_{\text{1:T}}).
\end{equation}
where $\text{Embedding}(\cdot)$ denotes a parameterized embedding function. To extract relevant information from the exogenous sequence, we introduce a learnable global context token $X_{\text{glob}}$, which is concatenated with the endogenous embedding along the temporal dimension to form an enhanced representation:
\begin{equation}
X_{\text{endo\_tok}} = \text{Concat}_{t}(X_{\text{endo}}, X_{\text{glob}}).
\end{equation}
where \(\text{Concat}_t(\cdot)\) denotes concatenation along the time dimension.

\subsection{Time-wise Gating Module (TGM)} TGM is designed to extract salient temporal patterns from historical endogenous sequences, as future values largely follow these dynamics. It employs an MLP to capture time-wise dependencies and map endogenous sequence information to the global token, generating time-wise gating weights that are subsequently passed through a sigmoid function to constrain them to a stable range, enabling selective feature filtering. TGM is formulated as:
\begin{multline}
[X'_{\text{endo}}, X'_{\text{glob}}] = 
\sigma\big(\text{Linear}_2\left(\phi\left(\text{Linear}_1(X_{\text{endo\_tok}})\right)\right)\big)\\
\odot X_{\text{endo\_tok}}.
\end{multline}
where $\text{Linear}_1$ and $\text{Linear}_2$ are learnable linear layers, 
$\phi(\cdot)$ (ReLU) and $\sigma(\cdot)$ (sigmoid) are the activation functions, 
and $\odot$ denotes element-wise multiplication.

\subsection{Variate-wise Gating Module (VGM)} VGM is designed to capture cross-variable information. It constructs dependency associations between endogenous variables' global tokens (obtained in the previous step) and the exogenous sequences. By doing so, it maps relevant information from exogenous variables into the endogenous global tokens, thereby optimizing the prediction of the endogenous variables. This step is formulated as:
\begin{equation}
X_{\text{exo\_tok}} = \text{Concat}_c(E_{\text{exo}}, X'_{\text{glob}}),
\end{equation}
\begin{equation}
[E'_{\text{exo}}, X''_{\text{glob}}] = \sigma\big(\text{Linear}_4(\phi(\text{Linear}_3(X_{\text{exo\_tok}})))\big) \odot X_{\text{exo\_tok}}.
\end{equation}
where $\text{Concat}_c(\cdot)$ denotes concatenation along the channel dimension.

\subsection{Prediction Head}
XLinear uses a fully connected layer as its final prediction head. It directly maps the integrated features to the target sequence by fusing the temporal characteristics of endogenous variables with key information extracted from exogenous variables. The computation follows:
\begin{equation}
    \hat{X}_{(T + 1):(T + S)} = \text{FC}\left(\text{Concat}_t\left(X'_{\text{endo}}, X''_{\text{glob}}\right)\right).
\end{equation}
where $\text{FC}(\cdot)$ denotes the fully connected layer.

\subsection{Loss Function} We employ Mean Square Error (MSE) as the loss function to quantify the deviation between predicted values and ground truth. Specifically, losses generated from individual endogenous channels are aggregated and then averaged across all $M$ endogenous variable sequences to obtain the overall loss. The calculation formula is as follows:
\begin{equation}
\mathcal{L} = \mathbb{E}_{\boldsymbol{X}} \frac{1}{M} \sum_{i=1}^{M} \left\| \hat{\boldsymbol{X}}_{(T+1):(T+S)}^{(i)} - \boldsymbol{X}_{(T+1):(T+S)}^{(i)} \right\|_2^2 .
\end{equation}
where $\mathbb{E}_{\boldsymbol{X}}$ denotes the expectation over time steps of $\boldsymbol{X}$, representing the average squared $L_2$-norm error between predictions and true values for from time steps $T+1$ to $T+S$.

\subsection{Multivariate Forecasting} In multivariate forecasting, each variable serves a dual purpose: it acts as an endogenous variable to be predicted and concurrently as an exogenous variable, providing variate-wise information to help predict other variables. To handle complex interaction, the TGM is shared across all variables, the VGM across features.

\begin{table*}[t]
    \centering
    \renewcommand{\arraystretch}{0.9}
    \scriptsize
    \begin{tabular}{lccccccc|cccccc}
        \toprule
        Datasets & Electricity & Weather & ETTh1 & ETTh2 & ETTm1 & ETTm2 & Traffic & Crop & DO$_{409215}$ & DO$_{425012}$ & GTD$_N$ & GTD$_S$\\
        \midrule
        Features & 321 & 21 & 7 & 7 & 7 & 7 & 862 & 8 & 7 & 7 & 16 & 16\\
        Timesteps & 26304 & 52696 & 17420 & 17420 & 69680 & 69680 & 17544 & 8914 & 103827 & 77336 & 38593 & 28727\\
        Granularity & 1 hour & 10 min & 1 hour & 1 hour & 15 min & 15 min & 1 hour & 5 min & 15 min & 15 min & 1 hour & 1 hour\\
        \bottomrule
    \end{tabular}
    \caption{Dataset characteristics: number of variables and sequence length, and temporal granularity (the first 7 are benchmarks).}
    \label{tab:datasets}
\end{table*}

\section{Experimental Results and Comparisons}

\subsection{Experimental Settings}

\textbf{Datasets.} We used 7 widely adopted datasets (ETT with 4 sub-datasets~\cite{zhou2021informer}, Weather, Electricity, Traffic~\cite{wu2021autoformer}), along with five datasets with exogenous inputs: one for \text{Crop} yield and four for environment monitoring. The Crop dataset came from Kaggle, two Dissolved Oxygen (DO) datasets, \(\text{DO}_{425012} \) and  \(\text{DO}_{409215}\), were from Darling River (u/s Weir 32) and the Murray River (Tocumwa), Australia~\citep{genova2025advancing}. Two water temperature datasets \( \text{GTD}_\text{N} \) and \( \text{GTD}_\text{S} \) were collected from the northern and southern monitoring sites of Grahamstown Dam, Australia, respectively\footnote{Please contact Warren Jin  for data sets~\citep{huang2025enhanced}.
}. Their metadata are presented in Table~\ref{tab:datasets}.

\noindent\textbf{Baselines.} We compare against SOTA models spanning multiple architectures. We include top-performing Transformer variants -- TimeXer~\cite{wang2024timexer}, which achieved the best results in the Long-term Forecasting Look-Back-96 scenario, and supports exogenous inputs, iTransformer~\cite{liu2023itransformer}, PatchTST~\cite{nie2022time}, and Crossformer~\cite{zhang2023crossformer}, as well as the CNN-based TimesNet~\cite{wu2022timesnet}. For linear-based models, we benchmark TimeMixer~\cite{wang2024timemixer}, RLinear~\cite{li2023revisiting}, TiDE~\cite{das2023long}, and DLinear~\cite{zeng2023transformers}. Finally, we compare with four leading GNN-based models~\cite{huang2023crossgnn,cai2024msgnet,wu2020connecting,yi2023fouriergnn}.

\noindent\textbf{Implementation Details.} We reproduced the baselines based on the official repositories provided by each model to ensure a fair comparison. For TiDE, we followed the implementation provided in the Time Series Library\footnote{https://github.com/thuml/Time-Series-Library} for replication.
All experiments were implemented using the PyTorch framework. The forecasting tasks on the PEMS07 and PEMS08 datasets (Table \ref{lab:pems}) were conducted on NVIDIA Quadro RTX 8000 (48 GB) GPU, while all other experiments were performed on NVIDIA Tesla V100-PCIE (32 GB) GPU. Following common practice in prior studies, we fixed the random seed to 2025 to ensure experimental reproducibility and control stochasticity across Python, NumPy, and PyTorch environments. For model optimization, the ADAM optimizer was employed. The initial learning rate was selected from the range $\{1 \times 10^{-4}, 2 \times 10^{-4}, 3 \times 10^{-4}, 5 \times 10^{-4}, 1 \times 10^{-3}\}$ based on the specific dataset used. The learning rate adjustment during training followed this rule: it remained constant for the first 3 epochs, and from the end of the 3rd epoch onwards, it was decayed by a factor of 0.9 for each subsequent epoch. Formally, the learning rate adjustment strategy can be expressed as:
$$\text{lr\_adjust}(e) = \begin{cases} 
\lambda_{\text{init}}, & e < 3 \\
\lambda_{\text{init}} \times 0.9^{(e-3)}, & e \geq 3 
\end{cases}$$
where $e$ denotes the current training epoch and $\lambda_{\text{init}}$ represents the initial learning rate. 

Considering the variations across different datasets, the batch size for training each dataset is selected from \(\{4, 8, 16, 32, 64, 128, 256\}\). During the training process, the number of training epochs is set to 30 in most cases, and an early stopping strategy is adopted. Detailed settings for each dataset can be found in the corresponding scripts.

Unless otherwise specified, the input length for all datasets was set to 96. The prediction horizon $S$ was selected from \{96,192,336,720\}, except for the Crop dataset, where it was chosen from \{12,24,36,48\}. For a time series of length $T$ (comprising $d_x$ endogenous variables and $d_z$ exogenous variables) with the batch size set as \(batch\_size\), the embedding layer projects both types of variable sequences into the same dimension $d_{model}$, where the length of the globally learnable tokens for endogenous variables is also $d_{model}$. In the temporal and channel gating modules, the hidden layer dimensions of the MLP are defined as $t_{ff}$  and $c_{ff}$, respectively, and the feature dimension remains unchanged after sigmoid processing. After capturing temporal dependencies, both the endogenous variable sequences and their global tokens have a dimension of $({batch\_size}, d_x, d_{model})$. To mitigate overfitting and enhance generalization, dropout regularization is introduced in the embedding layer, TGM, VGM, and prediction head, with dropout rates denoted as ${embed\_dropout}$, $t\_{dropout}$, $c\_{dropout}$, and ${head\_dropout}$, respectively. The model's prediction head maps the concatenated  endogenous sequence  tensor (shape $({batch\_size}, d_x, 2*d_{model})$) to the next $S$ time steps (shape (${batch\_size}, d_x, S))$ in a channel-independent manner.

\begin{table*}[t]
    \centering
    \setlength{\tabcolsep}{1mm}
    \renewcommand{\arraystretch}{0.85}
    \scriptsize
    \begin{tabular}{c|c|cc|cc|cc|cc|cc|cc|cc|cc|cc|cc}
        \toprule
        \multicolumn{2}{c}{\multirow{2}{*}{Models}} & \multicolumn{2}{c}{XLinear} & \multicolumn{2}{c}{TimeXer} & \multicolumn{2}{c}{TimeMixer} & \multicolumn{2}{c}{iTransformer} & \multicolumn{2}{c}{RLinear} & \multicolumn{2}{c}{PatchTST} & \multicolumn{2}{c}{Crossformer} & \multicolumn{2}{c}{TiDE} & \multicolumn{2}{c}{TimesNet} & \multicolumn{2}{c}{DLinear} \\
        \cmidrule(lr){3-4}\cmidrule(lr){5-6}\cmidrule(lr){7-8}\cmidrule(lr){9-10}\cmidrule(lr){11-12}\cmidrule(lr){13-14}\cmidrule(lr){15-16}\cmidrule(lr){17-18}\cmidrule(lr){19-20}\cmidrule(lr){21-22}
        \multicolumn{2}{c}{Metric} & MSE & MAE & MSE & MAE & MSE & MAE & MSE & MAE & MSE & MAE & MSE & MAE & MSE & MAE & MSE & MAE & MSE & MAE & MSE & MAE \\ 
        \toprule
        \multirow{5}{*}{\rotatebox{90}{ Electricity}} & 96 & \textbf{0.256} & \textbf{0.359} & \underline{0.261} & 0.366 & 0.275 & 0.384 & 0.299 & 0.403 & 0.433 & 0.480 & 0.339 & 0.412 & 0.265 & \underline{0.364} & 0.405 & 0.459 & 0.342 & 0.437 & 0.387 & 0.451 \\
        & 192 & \textbf{0.292} & \textbf{0.380} & 0.316 & 0.397 & \underline{0.294} & \underline{0.385} & 0.321 & 0.413 & 0.407 & 0.461 & 0.361 & 0.425 & 0.313 & 0.390 & 0.383 & 0.442 & 0.384 & 0.461 & 0.365 & 0.436 \\
        & 336 & \textbf{0.339} & \textbf{0.409} & 0.367 & 0.429 & \underline{0.344} & \underline{0.419} & 0.379 & 0.446 & 0.440 & 0.481 & 0.393 & 0.440 & 0.380 & 0.431 & 0.418 & 0.464 & 0.439 & 0.493 & 0.391 & 0.453 \\
        & 720 & \underline{0.385} & \underline{0.451} & \textbf{0.365} & \textbf{0.439} & 0.417 & 0.465 & 0.461 & 0.504 & 0.495 & 0.523 & 0.482 & 0.507 & 0.418 & 0.463 & 0.471 & 0.507 & 0.473 & 0.514 & 0.428 & 0.487 \\ 
        \midrule
        \multirow{5}{*}{\rotatebox{90}{ Weather}} & 96 & \textbf{0.001} & \underline{0.026} & \underline{0.001} & 0.027 & \underline{0.001} & 0.028 & \underline{0.001} & \underline{0.026} & \textbf{0.001} & \textbf{0.025} & \underline{0.001} & 0.027 & 0.004 & 0.048 & \textbf{0.001} & \textbf{0.025} & 0.002 & 0.029 & 0.006 & 0.062 \\
        & 192 & \textbf{0.001} & \underline{0.029} & \underline{0.002} & 0.030 & \underline{0.002} & 0.031 & \underline{0.002} & \underline{0.029} & \textbf{0.001} & \textbf{0.028} & \underline{0.002} & 0.030 & 0.005 & 0.053 & \textbf{0.001} & \textbf{0.028} & \underline{0.002} & 0.031 & 0.006 & 0.066 \\
        & 336 & \textbf{0.002} & \underline{0.030} & \underline{0.002} & 0.031 & \underline{0.002} & 0.031 & \underline{0.002} & 0.031 & \textbf{0.002} & \textbf{0.029} & \underline{0.002} & 0.032 & 0.004 & 0.051 & \textbf{0.002} & \textbf{0.029} & \underline{0.002} & 0.031 & 0.006 & 0.068 \\
        & 720 & \textbf{0.002} & \underline{0.034} & \underline{0.002} & 0.036 & \underline{0.002} & 0.036 & \underline{0.002} & 0.036 & \textbf{0.002} & \textbf{0.033} & \underline{0.002} & 0.036 & 0.007 & 0.067 & \textbf{0.002} & \textbf{0.033} & 0.381 & 0.368 & 0.007 & 0.070 \\ 
        \midrule
        \multirow{5}{*}{\rotatebox{90}{ ETTh1}} & 96 & \textbf{0.055} & \textbf{0.178} & 0.057 & \underline{0.181} & \underline{0.056} & \underline{0.181} & 0.057 & 0.183 & 0.059 & 0.185 & \textbf{0.055} & \textbf{0.178} & 0.133 & 0.297 & 0.059 & 0.184 & 0.059 & 0.188 & 0.065 & 0.188 \\
        & 192 & \textbf{0.071} & \textbf{0.202} & \textbf{0.071} & \underline{0.204} & \underline{0.072} & \underline{0.204} & 0.074 & 0.209 & 0.078 & 0.214 & \underline{0.072} & 0.206 & 0.232 & 0.409 & 0.078 & 0.214 & 0.080 & 0.217 & 0.088 & 0.222 \\
        & 336 & 0.084 & 0.226 & \textbf{0.080} & \textbf{0.223} & 0.088 & 0.230 & 0.084 & \textbf{0.223} & 0.093 & 0.240 & 0.087 & 0.231 & 0.244 & 0.423 & 0.093 & 0.240 & \underline{0.083} & \underline{0.224} & 0.110 & 0.257 \\
        & 720 & \textbf{0.083} & \textbf{0.227} & \underline{0.084} & \underline{0.229} & 0.086 & 0.230 & \underline{0.084} & \underline{0.229} & 0.106 & 0.256 & 0.098 & 0.247 & 0.530 & 0.660 & 0.104 & 0.255 & 0.083 & 0.231 & 0.202 & 0.371 \\ 
        \midrule
        \multirow{5}{*}{\rotatebox{90}{ ETTh2}} & 96 & \textbf{0.130} & \textbf{0.277} & \underline{0.132} & \underline{0.280} & 0.134 & 0.282 & 0.137 & 0.287 & 0.136 & 0.286 & 0.136 & 0.285 & 0.261 & 0.413 & 0.136 & 0.285 & 0.159 & 0.310 & 0.135 & 0.282 \\
        & 192 & \textbf{0.180} & \textbf{0.331} & \underline{0.181} & \underline{0.333} & 0.186 & 0.340 & 0.187 & 0.341 & 0.187 & 0.339 & 0.185 & 0.337 & 1.240 & 1.028 & 0.187 & 0.339 & 0.196 & 0.351 & 0.188 & 0.335 \\
        & 336 & \textbf{0.209} & \textbf{0.365} & 0.223 & 0.377 & 0.222 & 0.378 & 0.221 & 0.376 & 0.231 & 0.384 & \underline{0.217} & \underline{0.373} & 0.974 & 0.874 & 0.231 & 0.384 & 0.232 & 0.385 & 0.238 & 0.385 \\
        & 720 & \textbf{0.217} & \textbf{0.373} & \underline{0.220} & \underline{0.376} & 0.223 & 0.379 & 0.253 & 0.403 & 0.267 & 0.417 & 0.229 & 0.384 & 1.633 & 1.177 & 0.267 & 0.417 & 0.254 & 0.403 & 0.336 & 0.475 \\ 
        \midrule
        \multirow{5}{*}{\rotatebox{90}{ ETTm1}} & 96 & \textbf{0.028} & \textbf{0.125} & \textbf{0.028} & \textbf{0.125} & \underline{0.029} & \underline{0.126} & \underline{0.029} & 0.128 & 0.030 & 0.129 & \underline{0.029} & \underline{0.126} & 0.171 & 0.355 & 0.030 & 0.129 & \underline{0.029} & 0.128 & 0.034 & 0.135 \\
        & 192 & \textbf{0.043} & \textbf{0.158} & \textbf{0.043} & \textbf{0.158} & \underline{0.044} & \underline{0.160} & 0.045 & 0.163 & \underline{0.044} & \underline{0.160} & 0.045 & \underline{0.160} & 0.293 & 0.474 & \underline{0.044} & \underline{0.160} & \underline{0.044} & \underline{0.160} & 0.055 & 0.173 \\
        & 336 & \textbf{0.057} & \textbf{0.183} & \underline{0.058} & 0.185 & 0.059 & 0.186 & 0.060 & 0.190 & \textbf{0.057} & \underline{0.184} & \underline{0.058} & \underline{0.184} & 0.330 & 0.503 & \textbf{0.057} & \underline{0.184} & 0.061 & 0.190 & 0.078 & 0.210 \\
        & 720 & \textbf{0.079} & \textbf{0.216} & \textbf{0.079} & \underline{0.217} & 0.082 & 0.218 & \textbf{0.079} & 0.218 & \underline{0.080} & \underline{0.217} & 0.082 & 0.221 & 0.852 & 0.861 & \underline{0.080} & \underline{0.217} & 0.083 & 0.223 & 0.098 & 0.234 \\ 
        \midrule
        \multirow{5}{*}{\rotatebox{90}{ ETTm2}} & 96 & \textbf{0.064} & \textbf{0.182} & \underline{0.067} & \underline{0.188} & 0.068 & 0.189 & 0.071 & 0.194 & 0.074 & 0.199 & 0.068 & 0.188 & 0.149 & 0.309 & 0.073 & 0.199 & 0.073 & 0.200 & 0.072 & 0.195 \\
        & 192 & \textbf{0.098} & \textbf{0.233} & 0.101 & 0.236 & \underline{0.100} & \underline{0.234} & 0.108 & 0.247 & 0.104 & 0.241 & \underline{0.100} & 0.236 & 0.686 & 0.740 & 0.104 & 0.241 & 0.106 & 0.247 & 0.105 & 0.240 \\ 
        & 336 & \underline{0.129} & \underline{0.274} & 0.130 & 0.275 & 0.132 & 0.278 & 0.140 & 0.288 & 0.131 & 0.276 & \textbf{0.128} & \textbf{0.271} & 0.546 & 0.602 & 0.131 & 0.276 & 0.150 & 0.296 & 0.136 & 0.280 \\
        & 720 & \textbf{0.179} & \textbf{0.329} & 0.182 & \underline{0.332} & 0.183 & \underline{0.332} & 0.188 & 0.340 & \underline{0.180} & \textbf{0.329} & 0.185 & 0.335 & 2.524 & 1.424 & \underline{0.180} & \textbf{0.329} & 0.186 & 0.338 & 0.191 & 0.335 \\
        \midrule
        \multirow{5}{*}{\rotatebox{90}{ Traffic}} & 96 & \textbf{0.139} & \textbf{0.216} & \underline{0.151} & \underline{0.224} & 0.170 & 0.256 & 0.156 & 0.236 & 0.350 & 0.431 & 0.176 & 0.253 & 0.154 & 0.230 & 0.350 & 0.430 & 0.154 & 0.249 & 0.268 & 0.351 \\
        & 192 & \textbf{0.140} & \textbf{0.214} & \underline{0.152} & \underline{0.229} & 0.162 & 0.248 & 0.156 & 0.237 & 0.314 & 0.404 & 0.162 & 0.243 & 0.180 & 0.256 & 0.230 & 0.315 & 0.164 & 0.255 & 0.302 & 0.387 \\
        & 336 & \textbf{0.143} & \textbf{0.221} & \underline{0.150} & \underline{0.232} & 0.159 & 0.251 & 0.154 & 0.243 & 0.305 & 0.399 & 0.164 & 0.248 & {-} & {-} & 0.220 & 0.208 & 0.167 & 0.259 & 0.298 & 0.384 \\
        & 720 & \textbf{0.165} & \textbf{0.245} & \underline{0.172} & \underline{0.253} & 0.186 & 0.274 & 0.177 & 0.268 & 0.328 & 0.415 & 0.189 & 0.267 & {-} & {-} & 0.243 & 0.329 & 0.197 & 0.292 & 0.340 & 0.416 \\ 
        \midrule
        \multicolumn{2}{c|}{1st} & 25 & 21 & 6 & 4 & 0 & 0 & 1 & 1 & 5 & 5 & 2 & 2 & 0 & 0 & 5 & 5 & 0 & 0 & 0 & 0 \\ 
        \bottomrule
    \end{tabular}
    \caption{Results of univariate forecasting with exogenous inputs. ``-'' denotes the occurrence of an Out of Memory (OOM) exception during model execution. The best results are in \textbf{bold} and the second best are \underline{underlined}. All baseline results except those of TimeMixer are reported in~\cite{wang2024timexer}.}
    \label{tab:univariate_forcasting}
\end{table*}

\noindent\textbf{Evaluation Metrics.}
For the seven public datasets, we follow the evaluation conventions in existing studies and adopt MSE and Mean Absolute Error (MAE) as evaluation metrics. For the Crop dataset and environmental datasets, we use  additional  evaluation metrics such as Nash-Sutcliffe Efficiency (NSE), Kling-Gupta Efficiency (KGE), and Mean Absolute Percentage Error (MAPE) which are widely employed in environmental sciences, enabling a more holistic assessment of forecast model performance~\cite{gupta2009decomposition}.
They are defined as:
$  \text{NSE} = 1 - \frac{\sum_{i=1}^{S} (y_i - \hat{y}_i)^2}{\sum_{i=1}^{S} (y_i - \bar{y})^2}
$
where \(y_i\) denotes the observed value, \(\hat{y}_i\) represents the predicted value, \(\bar{y}\) is the mean of observed values, and \(n\) is the number of samples; 
$
    \text{KGE} = 1 - \sqrt{(r - 1)^2 + (\alpha - 1)^2 + (\beta - 1)^2}
$
where \(r\) is the Pearson correlation coefficient between observed and predicted values, variability ratio \(\alpha\) is the ratio of the standard deviations of the predicted and the observed values, and bias ratio \(\beta\)  indicates the ratio of the means of the predicted and the observed values; 
$
    \text{MAPE} = \frac{1}{n} \sum_{i=1}^{n} \left| \frac{y_i - \hat{y}_i}{y_i} \right| \times 100\%
$.
Higher NSE and KGE values indicate better performance, while lower values are preferred for the other metrics.

\begin{table*}[t]
    \centering
    \setlength{\tabcolsep}{1mm}
    \renewcommand{\arraystretch}{0.85}
    \scriptsize
    \begin{tabular}{c|c|ccccc|ccccc|ccccc|ccccc}
        \toprule
        \multicolumn{2}{c}{\multirow{2}{*}{{Models}}} & \multicolumn{5}{c}{{XLinear}} & \multicolumn{5}{c}{{TimeXer}} & \multicolumn{5}{c}{{iTransformer}} & \multicolumn{5}{c}{{PatchTST}}  \\
        \cmidrule(lr){3-7}\cmidrule(lr){8-12}\cmidrule(lr){13-17}\cmidrule(lr){18-22}
        \multicolumn{2}{c}{{Metric}}   & {MSE} & {MAE} & {NSE}    & {KGE}   & {MAPE} & {MSE} & {MAE} & {NSE}    & {KGE}   & {MAPE} & {MSE} & {MAE} & {NSE}    & {KGE}   & {MAPE} & {MSE} & {MAE} & {NSE}    & {KGE}   & {MAPE} \\ 
        \toprule[1.2pt]
        \multirow{5}{*}{{\rotatebox{90}{DO$_{425012}$}}} & {96} & {\textbf{0.113}} & {\textbf{0.207}} & {\textbf{0.862}} & {\textbf{0.930}} & {\textbf{0.451}} & {\underline{0.116}} & {\underline{0.212}} & {\underline{0.858}} & {\underline{0.928}}  & {0.460} & {0.117} & {\underline{0.212}} & {0.857} & {0.927} & {\underline{0.452}} & {0.117} & {\underline{0.212}} & {\underline{0.858}} & {0.925} & {0.466} \\
        & {192} & {\textbf{0.212}} & {\textbf{0.288}} & {\textbf{0.741}} & {\textbf{0.869}} & {\underline{0.653}} & {0.215} & {\textbf{0.288}} & {0.738} & {0.863} & {\textbf{0.650}} & {\underline{0.214}} & \underline{0.289} & {\underline{0.740}} & {\underline{0.867}} & \underline{0.653} & {\underline{0.214}} & {0.290} & {0.739} & {0.865} & {0.662}\\
        & {336} & {\underline{0.340}} & {\textbf{0.379}} & {\underline{0.586}} & {\textbf{0.789}} & {0.935} & {0.341} & {\textbf{0.379}} & {0.585} & {\underline{0.786}} & {\textbf{0.924}} & {\textbf{0.337}} & {\underline{0.380}} & {\textbf{0.590}} & {\textbf{0.789}} & {\underline{0.930}} & {0.345} & {0.382} & {0.580} & {0.783} & {0.948}\\
        & {720} & {\textbf{0.555}} & {\textbf{0.531}} & {\textbf{0.329}} & {\underline{0.649}} & {\underline{1.526}} & {0.561} & {0.537} & {0.322} & {\textbf{0.653}} & {{1.540}} & {\underline{0.556} }& {\textbf{0.531}} & {\underline{0.328}} & {0.648} & {\textbf{1.512}} & {0.562} & {\underline{0.536}} & {0.320} & {0.645} & {1.552}\\ 
        \midrule
        \multirow{5}{*}{{\rotatebox{90}{DO$_{409215}$}}} & {96} & {\textbf{0.003}} & {\textbf{0.037}} & {\textbf{0.986}} & {\textbf{0.992}} & {\textbf{1.014}} & {\textbf{0.003}} & {\underline{0.038}} & {\underline{0.985}} & {0.964} & {\underline{1.022}} & {\textbf{0.003}} & {\underline{0.038}} & {0.984} & {0.953} & {1.043} & {\textbf{0.003}} & {0.041} & {0.983} & {\underline{0.976}}  & {{1.027}}\\
        & {192} & {\textbf{0.006}} & {\textbf{0.054}} & {\textbf{0.970}} & {\textbf{0.982}} & {\underline{1.413}} & {\textbf{0.006}} & {\textbf{0.054}} & {\underline{0.969}} & {0.930} & {1.418} & {\textbf{0.006}} & {\underline{0.055}} & {0.968} & {0.926} & {1.428} & {\underline{0.007}} & {0.059} & {0.965} & {\underline{0.957}} & {\textbf{1.404}} \\
        & {336} & {\textbf{0.011}} & {\textbf{0.077}} & {\textbf{0.942}} & {\textbf{0.967}} & {\textbf{1.683}} & {\underline{0.012}} & {\underline{0.079}} & {\underline{0.939}} & {0.909} & {1.774} & {\underline{0.012}} & {\underline{0.079}} & {0.938} & {0.882} & {\underline{1.738}} & {0.013} & {0.081} & {0.936} & {\underline{0.953}} & {1.748} \\
        & {720} & {\textbf{0.023}} & {\textbf{0.111}} & {\textbf{0.880}} & {\textbf{0.913}} & {\textbf{2.397}} & {0.027} & {\underline{0.118}} & {{0.861}} & {0.844} & {\underline{2.603}} & {0.027} & {0.121} & {0.859} & {0.818} & {2.854} & {\underline{0.026}} & {0.119} & {\underline{0.862}} & {\underline{0.904}} & {2.820} \\ 
        \midrule
        \multirow{5}{*}{{\rotatebox{90}{GTD$_{N}$}}} & {96} & {\textbf{0.008}} & {\textbf{0.062}} & {\textbf{0.991}} & {\underline{0.864}} & {\textbf{0.455}} & {\underline{0.009}} & {0.064} & {\underline{0.990}} & {0.781} & {\underline{0.458}} & {\underline{0.009}} & {\underline{0.063}} & {\underline{0.990}} & {0.851} & {\textbf{0.455}} & {0.010} & {0.068} & {0.989} & {\textbf{0.898}} & {0.481} \\
        & {192} & {\textbf{0.018}} & {\textbf{0.092}} & {\textbf{0.980}} & {\textbf{0.861}} & {\textbf{0.726}} & \underline{0.019} & {0.096} & {0.978} & {0.775} & {0.790} & {\textbf{0.018}} & {\underline{0.094}} & {\underline{0.979}} & {\underline{0.792}} & {0.751} & {0.020} & {0.100} & {0.977} & {0.747} & {\underline{0.740}} \\
        & {336} & {\textbf{0.033}} & {\textbf{0.127}} & {\textbf{0.962}} & {\textbf{0.824}} & {\underline{0.864}} & {{0.036}} & {0.133} & {0.958} & {0.741} & {0.917} & {\underline{0.035}} & {\underline{0.130}} & {\underline{0.959}} & {\underline{0.801}} & {\textbf{0.855}} & {0.037} & {0.138} & {0.957} & {0.661} & {0.933} \\
        & {720} & {\underline{0.093}} & {\underline{0.221}} & {\underline{0.889}} & {\underline{0.567}} & {\underline{1.651}} & {{0.098}} & {0.226} & {0.883} & {0.441} & {1.658} & {\textbf{0.062}} & {\textbf{0.180}} & {\textbf{0.926}} & {\textbf{0.749}} & {\textbf{1.225}} & {0.104} & {0.239} & {0.877} & {0.362} & {1.837} \\ 
        \midrule
        \multirow{5}{*}{{\rotatebox{90}{GTD$_{S}$}}} & {96}  & {\textbf{0.012}} & {\textbf{0.077}} & {\textbf{0.971}} & {\underline{0.983}} & {\textbf{0.554}} & {\underline{0.013}} & {\underline{0.079}} & \underline{0.970} & {0.978} & {\underline{0.558}} & {\underline{0.013}} & {0.081} & {0.969} & {\textbf{0.984}} & {0.602} & {\underline{0.013}} & {0.082} & {0.968} & {0.982} & {0.591} \\
        & {192} & {\textbf{0.025}} & {\textbf{0.112}} & {\textbf{0.936}} & {\underline{0.965}} & {\textbf{0.729}} & {\textbf{0.025}} & {\underline{0.114}} & {\textbf{0.936}} & {\textbf{0.966}} & {\underline{0.748}} & {\underline{0.027}} & {0.118} & {\underline{0.933}} & {0.964} & {0.786} & {\underline{0.027}} & {0.118} & {0.931} & {0.961} & {0.755} \\
        & {336} & {\textbf{0.048}} & {\textbf{0.159}} & {\underline{0.869}} & {{0.926}} & {\underline{0.941}} & {\underline{0.049}} & {\textbf{0.159}} & {\textbf{0.870}} & {\underline{0.928}} & {\textbf{0.921}} & {0.051} & {0.164} & {0.863} & {\textbf{0.930}} & {1.002} & {0.050} & \underline{0.162} & {0.864} & {0.923} & {0.945} \\
        & {720} & {\underline{0.124}} & {\underline{0.260}} & {0.621} & {\underline{0.795}} & {1.550} & {\textbf{0.121}} & {\textbf{0.258}} & {\textbf{0.634}} & {\textbf{0.816}} & {\underline{1.462}} & {0.125} & {0.268} & {\underline{0.622}} & {0.782} & {\textbf{1.370}} & {0.131} & {0.268} & {0.600} & {0.785} & {1.631} \\ 
        \midrule
        \multirow{5}{*}{{\rotatebox{90}{Crop}}} & {12}  & 0.147 & \textbf{0.236} & \underline{0.529} & \underline{0.714} & 0.982 & \textbf{0.143} & 0.245 & \textbf{0.541} & 0.689 & \underline{0.940} & 0.153 & 0.251 & 0.510 & \textbf{0.719} & 0.991 & \underline{0.144} & \underline{0.244} & \underline{0.529} & 0.705 & \textbf{0.920} \\
        & {24} & \textbf{0.186} & \underline{0.284} & \textbf{0.408} & \underline{0.612} & 1.071 & 0.190 & 0.296 & 0.392 & 0.606 & \underline{1.019} & 0.208 & 0.307 & 0.336 & 0.607 & 1.131 & \textbf{0.186} & \textbf{0.280} & \underline{0.394} & \textbf{0.616} & \textbf{1.018} \\
        & {36} & \textbf{0.219} & \underline{0.311} & \textbf{0.303} & \underline{0.542} & 1.137 & 0.224 & 0.327 & \underline{0.287} & \textbf{0.547} & \underline{1.092} & 0.247 & 0.343 & 0.212 & 0.448 & 1.205 & \underline{0.220} & \textbf{0.310} & \underline{0.287} & 0.539 & \textbf{1.077} \\
        & {48} & \textbf{0.245} & \textbf{0.336} & \textbf{0.220} & 0.433 & \underline{1.175} & \underline{0.251} & 0.356 & \underline{0.203} & \textbf{0.485} & 1.185 & 0.274 & 0.367 & 0.129 & 0.388 & 1.245 & 0.252 & \underline{0.338} & 0.188 & \underline{0.462} & \textbf{1.100} \\ 
        \midrule
        \multicolumn{2}{c|}{{{$1^{\text{st}}$ Count}}} & {16} & {16} & {15} & {9} & {8} & {5} & {5} & {4} & {5} & {3} & {5} & {2} & {2} & {5} & {5} & {2} & {2} & {0} & {2} & {5} \\ \bottomrule
    \end{tabular}
    \caption{Results of single endogenous variable prediction based on exogenous variables for each model on environmental datasets. The best results are in \textbf{bold} and the second best are \underline{underlined}.}
    \label{tab:environment results}
\end{table*}

\begin{table*}[t]
    \centering
    \setlength{\tabcolsep}{1mm}
    \renewcommand{\arraystretch}{0.85}
    \scriptsize
    \begin{tabular}{c|c|cc|cc|cc|cc|cc|cc|cc|cc|cc|cc}
        \toprule
         \multicolumn{2}{c}{\multirow{2}{*}{{Models}}}  & \multicolumn{2}{c}{{XLinear}} & \multicolumn{2}{c}{{TimeXer}} & \multicolumn{2}{c}{{TimeMixer}} & \multicolumn{2}{c}{{iTransformer}} & \multicolumn{2}{c}{{RLinear}} & \multicolumn{2}{c}{{PatchTST}} & \multicolumn{2}{c}{{Crossformer}} & \multicolumn{2}{c}{{TiDE}} & \multicolumn{2}{c}{{TimesNet}} & \multicolumn{2}{c}{{DLinear}} \\ 
         \cmidrule(lr){3-4}\cmidrule(lr){5-6}\cmidrule(lr){7-8}\cmidrule(lr){9-10}\cmidrule(lr){11-12}\cmidrule(lr){13-14}\cmidrule(lr){15-16}\cmidrule(lr){17-18}\cmidrule(lr){19-20}\cmidrule(lr){21-22}
         \multicolumn{2}{c}{{Metric}} & {MSE} & {MAE} & {MSE} & {MAE} & {MSE} & {MAE} & {MSE}    & {MAE} & {MSE} & {MAE} & {MSE} & {MAE}  & {MSE} & {MAE} & {MSE} & {MAE} & {MSE} & {MAE} & {MSE} & {MAE}\\

         \toprule
        \multirow{5}{*}{\rotatebox{90}{Electricity}} 
        &  {96} & {\textbf{0.138}} & {\textbf{0.237}} & {\underline{0.140}} & {{0.242}} & 0.153 & 0.247 & {{0.148}} & {\underline{0.240}} & {0.201} & {0.281} & {0.195} & {0.285} & {0.219} & {0.314} & {0.237} & {0.329} & {0.168} & {0.272} &{0.197} &{0.282} \\ 
        & {192} & {\textbf{0.156}} & {\textbf{0.251}} & {\underline{0.157}} & {{0.256}} & 0.166 & 0.256 & {0.162} & {\underline{0.253}} & {0.201} & {0.283} & {0.199} & {0.289} & {0.231} & {0.322} & {0.236} & {0.330} &{{0.184}} &{0.289} &{0.196} &{{0.285}} \\ 
        & {336} & {\textbf{0.171}} & {\textbf{0.267}} & {\underline{0.176}} & {0.275} & 0.185 & 0.277 &  {{0.178}} & {\underline{0.269}} & {0.215} & {0.298} & {0.215} & {0.305} & {0.246} & {0.337} & {0.249} & {0.344} & {0.198} & {0.300} &{0.209} & {0.301} \\
        & {720} & {\textbf{0.206}} & {\textbf{0.297}} & {\underline{0.211}} & {\underline{0.306}} & 0.225 & 0.310 & {0.225} & {0.317} & {0.257} & {0.331} & {0.256} & {0.337} & {0.280} & {0.363} & {0.284} & {0.373} & {{0.220}} & {0.320} & {0.245} & {0.333} \\ 
        \midrule
        \multirow{5}{*}{\rotatebox{90}{{Weather}}} 
        & {96} & {\textbf{0.149}} & {\textbf{0.198}} & {\underline{0.157}} & {\underline{0.205}} & 0.163 & 0.209 & {0.174} & {0.214} & {0.192} & {0.232} & {0.177} & {0.218} & {0.158} & {0.230}  & {0.202} & {0.261} & {0.172} & {0.220} & {0.196} & {0.255} \\ 
        & {192} & {\textbf{0.201}} & {\textbf{0.244}} & {\underline{0.204}} & {\underline{0.247}} & 0.208 & 0.250 &  {0.221} & {0.254} & {0.240} & {0.271} & {0.225} & {0.259} & {{0.206}} & {0.277} & {0.242} & {0.298} & {0.219} & {0.261} & {0.237} & {0.296} \\ 
        & {336} & {\underline{0.259}} & {\underline{0.288}} & {{0.261}} & {{0.290}} & \textbf{0.251} & \textbf{0.287} & {0.278} & {0.296} & {0.292} & {0.307} & {0.278} & {0.297} & {0.272} & {0.335} & {0.287} & {0.335} & {0.280} & {0.306} & {0.283} & {0.335} \\ 
        & {720} & {\textbf{0.339}} & {\textbf{0.340}} & {\underline{0.340}} & {\underline{0.341}} & \textbf{0.339} & \underline{0.341} & {0.358} & {0.349} & {0.364} & {0.353} & {0.354} & {0.348} & {0.398} & {0.418} & {0.351} & {0.386} & {0.365} & {0.359} & {{0.345}} & {0.381} \\ 
        \midrule
        \multirow{5}{*}{\rotatebox{90}{{ETTh1}}}
        & {96} & {\textbf{0.369}} & {\textbf{0.393}} & {{0.382}} & {0.403} & \underline{0.375} & 0.400 & {0.386} & {0.405} & {0.386} & {\underline{0.395}} & {0.414} & {0.419} & {0.423} & {0.448} & {0.479} & {0.464}  & {0.384} & {0.402} & {0.386} & {0.400} \\
        & {192} & {\textbf{0.421}} & {\underline{0.423}} & {\underline{0.429}} & {0.435} & \underline{0.429} & \textbf{0.421} & {0.441} & {0.436} & {0.437} & {{0.424}} & {0.460} & {0.445} & {0.471} & {0.474} & {0.525} & {0.492} & {0.436} & {0.429} & {0.437} & {0.432} \\ 
        & {336} & {\textbf{0.455}} & {\textbf{0.440}} & {\underline{0.468}} & {{0.448}} & 0.484 & 0.458 & {0.487} & {0.458} & {0.479} & {\underline{0.446}} & {0.501} & {0.466} & {0.570} & {0.546} & {0.565} & {0.515} & {0.491} & {0.469} & {0.481}  & {0.459} \\
        & {720} & {\textbf{0.453}} & {\textbf{0.456}} & {\underline{0.469}} & {\underline{0.461}} & 0.498 & 0.482 & {0.503} & {{0.491}} & {{0.481}} & {{0.470}} & {0.500} & {0.488} & {0.653} & {0.621} & {0.594} & {0.558} & {0.521} & {0.500} & {0.519} & {0.516}  \\
        \midrule
        \multirow{5}{*}{\rotatebox{90}{{ETTh2}}}
        & {96} & \textbf{{0.286}} & {\textbf{0.337}} & {\textbf{0.286}} & {\underline{0.338}} & 0.289 & \underline{0.341} & {0.297} & {0.349} & {\underline{0.288}} & {\underline{0.338}} & {0.302} & {0.348} & {0.745} & {0.584} & {0.400} & {0.440} & {0.340} & {0.374} & {0.333} & {0.387} \\ 
        & {192} & {\textbf{0.363}} & {\textbf{0.388}} & {\textbf{0.363}} & {\underline{0.389}} & \underline{0.372} & 0.392 & {0.380} & {0.400} & {0.374} & {\underline{0.390}} & {0.388} & {0.400} & {0.877} & {0.656} & {0.528} & {0.509} & {0.402} & {0.414} & {0.477} & {0.476} \\
        & {336} & {\textbf{0.378}} & {\textbf{0.407}} & {{0.414}} & {{0.423}} & \underline{0.386} & \underline{0.414} & {0.428} & {0.432} & {{0.415}} & {0.426} & {0.426} & {0.433} & {1.043} & {0.731} & {0.643} & {0.571}  & {0.452} & {0.452} & {0.594} & {0.541} \\
        & {720} & {\textbf{0.408}} & {\textbf{0.431}} & {\textbf{0.408}} & {\underline{0.432}} & \underline{0.412} & 0.434 & {0.427} & {0.445} & {{0.420}} & {{0.440}} & {0.431} & {0.446} & {1.104} & {0.763} & {0.874} & {0.679} & {0.462} & {0.468} & {0.831} & {0.657} \\
        \midrule
        \multirow{5}{*}{\rotatebox{90}{{ETTm1}}}
        & {96} & {\textbf{0.311}} & {\textbf{0.351}} & {\underline{0.318}} & {\underline{0.356}} & 0.320 & 0.357 & {0.334} & {0.368} & {0.355} & {0.376} & {0.329} & {0.367} & {0.404} & {0.426} & {0.364} & {0.387} & {0.338} & {0.375} & {0.345} & {0.372} \\
        & {192} & {\textbf{0.353}} & {\textbf{0.376}} & {{0.362}} & {{0.383}} & \underline{0.361} & \underline{0.381} & {0.387} & {0.391} & {0.391} & {0.392} & {0.367} & {0.385} & {0.450} & {0.451} & {0.398} & {0.404} & {0.374} & {0.387} & {0.380} & {0.389} \\
        & {336} & {\textbf{0.382}} & {\textbf{0.398}} & {{0.395}} & {{0.407}} & \underline{0.390} & \underline{0.404} & {0.426} & {0.420} & {0.424} & {0.415} & {0.399} & {0.410} & {0.532} & {0.515} & {0.428} & {0.425} & {0.410} & {0.411}  & {0.413} & {0.413} \\
        & {720} & {\textbf{0.444}} & {\textbf{0.436}} & {\underline{0.452}} & {{0.441}} & 0.454 & 0.441 & {0.491} & {0.459} & {0.487} & {0.450} & {0.454} & {\underline{0.439}} & {0.666} & {0.589} & {0.487} & {0.461} & {0.478} & {0.450} & {0.474} & {0.453} \\
        \midrule
        \multirow{5}{*}{\rotatebox{90}{{ETTm2}}}
        & {96} & {\textbf{0.167}} & {\textbf{0.250}} & {\underline{0.171}} & {\underline{0.256}} & 0.175 & 0.258 & {0.180} & {0.264} & {0.182} & {0.265} & {0.175} & {0.259} & {0.287} & {0.366} & {0.207} & {0.305} & {0.187} & {0.267} & {0.193} & {0.292}  \\ 
        & {192} & {\textbf{0.233}} & {\textbf{0.293}} & {\underline{0.237}} & {\underline{0.299}} & \underline{0.237} & \underline{0.299} & {0.250} & {0.309} & {0.246} & {0.304} & {0.241} & {0.302} & {0.414} & {0.492} & {0.290} & {0.364} & {0.249} & {0.309} & {0.284} & {0.362} \\ 
        & {336} & {\textbf{0.290}} & {\textbf{0.332}} & {\underline{0.296}} & {\underline{0.338}} & 0.298 & 0.340 & {0.311} & {0.348} & {0.307} & {0.342} & {0.305} & {0.343} & {0.597} & {0.542} & {0.377} & {0.422} & {0.321} & {0.351} & {0.369} & {0.427} \\ 
        & {720} & {\textbf{0.388}} & {\textbf{0.390}} & {{0.392}} & {\underline{0.394}} & \underline{0.391} & 0.396 & {0.412} & {0.407} & {0.407} & {0.398} & {0.402} & {0.400} & {1.730} & {1.042} & {0.558} & {0.524} & {0.408} & {0.403} & {0.554} & {0.522} \\
        \midrule
        \multirow{5}{*}{\rotatebox{90}{{Traffic}}} 
        & {96} & {0.432} & {0.281} & {\underline{0.428}} & {\underline{0.271}} & 0.462 & 0.285 & {\textbf{0.395}} & {\textbf{0.268}} & {0.649} & {0.389} & {0.462} & {0.295} & {0.522} & {0.290} & {0.805} & {0.493} & {0.593} & {0.321} & {0.650} & {0.396} \\ 
        & {192} & {0.449} & {0.289} & {\underline{0.448}} & {\underline{0.282}} & 0.473 & 0.296 & {\textbf{0.417}} & {\textbf{0.276}} & {0.601} & {0.366} & {0.466} & {0.296} & {0.530} & {0.293} & {0.756} & {0.474} & {0.617} & {0.336} & {0.598} & {0.370}  \\
        & {336} & {\underline{0.465}} & {0.295} & {0.473} & {\underline{0.289}} & 0.498 & 0.296 & {\textbf{0.433}} & {\textbf{0.283}} & {0.609} & {0.369} & {0.482} & {0.304} & {0.558} & {0.305} & {0.762} & {0.477} & {0.629} & {0.336} & {0.605} & {0.373} \\
        & {720} & {{0.507}} & {0.315} & {0.516} & {\underline{0.307}} & \underline{0.506} & 0.313 & {\textbf{0.467}} & {\textbf{0.302}} & {0.647} & {0.387} & {0.514} & {0.322} & {0.589} & {0.328} & {0.719} & {0.449} & {0.640} & {0.350} & {0.645} & {0.394} \\
        \midrule
        \multicolumn{2}{c|}{{{$1^{\text{st}}$ Count}}} & {23} & {22} & {3} & {0} & {2} & {2} & {4} & {4} & {0} & {0} & {0} & {0} & {0} & {0} & {0} & {0} & {0} & {0} & {0} & {0}
        \\ \bottomrule
    \end{tabular}
    \caption{Multivariate time-series forecasting results. The best results are in \textbf{bold} and the second best are \underline{underlined}. The results of TimeMixer are reported in~\cite{wang2024timemixer}, while those of other baselines are reported in~\cite{wang2024timexer}.}
    \label{tab:multi_forcasting}
\end{table*}

\subsection{Main Results}

\noindent\textbf{Univariate Forecast with Exogenous inputs.} In each dataset, for univariate forecast performance assessment, the final variable is designated as the endogenous variable, while the remaining variables are treated as exogenous variables, following common practice~\cite{wang2024timexer}. For the 5 environmental datasets, their last variable consistently serves as the endogenous one. In the Crop dataset, yield is driven by soil moisture, air temperature, humidity, and etc. In the water quality dataset, DO$_{425012}$ and DO$_{409215}$, dissolved oxygen content is guided by water temperature, mean discharge rate, and mean water level, among other variables. In the GTD$_N$ and GTD$_S$ datasets, the bottom water temperature at 9m depth is affected by shortwave radiation, air temperature, cloud cover, wind speed and direction, etc. 

The forecast MSE and MAE results for the 7 benchmarks are listed in Table \ref{tab:univariate_forcasting}. XLinear achieves the lowest errors in 89.3\% (25/28) of experimental configurations for MSE, and 75.0\% for MAE. It consistently ranks among the top two models, except at the 336-step lead time for ETTh1 -- outperforming strong competitors such as TimeXer and TimeMixer in most cases. In addition to its accuracy, XLinear also demonstrates high efficiency. 
As illustrated in Fig \ref{fig:effience}, under relatively consistent parameter settings, among all baseline models, XLinear achieves at least 39.3\% faster training, 13.9\% lower memory usage, and 2.8\% lower MSE than the others—except DLinear and RLinear, whose MSEs are around 23.5\% and 39.6\% higher than XLinear, respectively.

We further evaluate XLinear on the five real-world applications. Due to space constraints, we compare it with three representative models, as listed in Table \ref{tab:environment results}. XLinear achieves the lowest MSE and MAE in over 75\% of the cases. For the composite metric NSE, it ranks first in 75\% of the rows, and second in 20\%. Additionally, XLinear reaches the ``Good" performance level (NSE $>$ 0.65, per~\cite{moriasi2007model}) in 65\% of the cases. For the KGE and MAPE metrics, XLinear ranks among the top two models in 90\%  and 70\%  of the cases, respectively. Overall, XLinear demonstrates consistent advantages across multiple evaluation metrics, confirming its effectiveness in real-world forecasting scenarios.

\begin{figure}[t]
    \centering
    \begin{tabular}{ccc}
        \includegraphics[width=0.28\columnwidth]{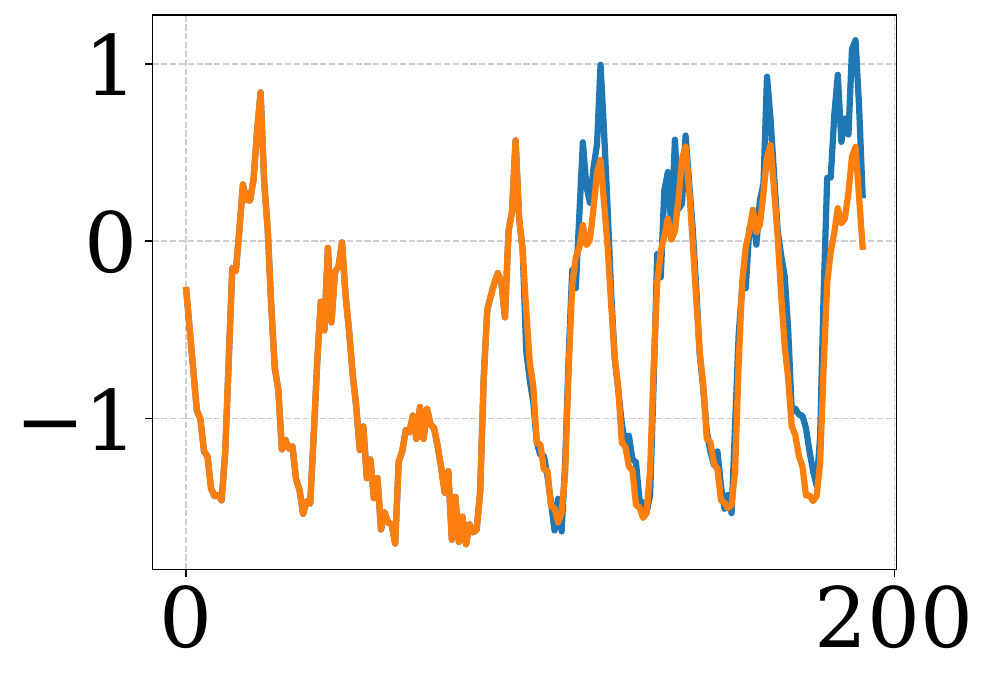} &
        \includegraphics[width=0.28\columnwidth]{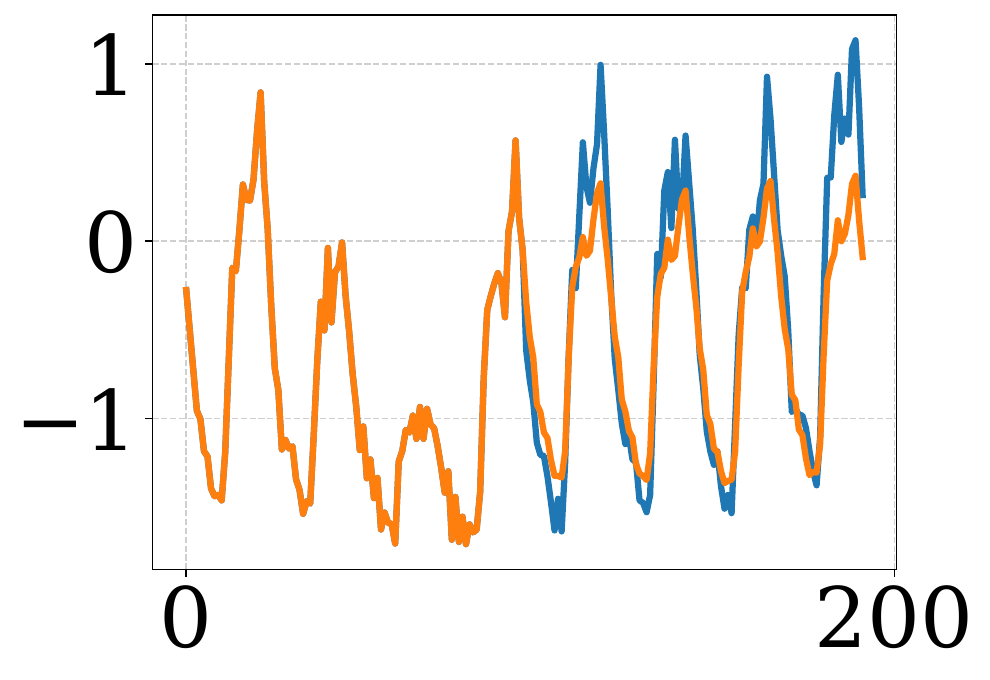} &
        \includegraphics[width=0.28\columnwidth]{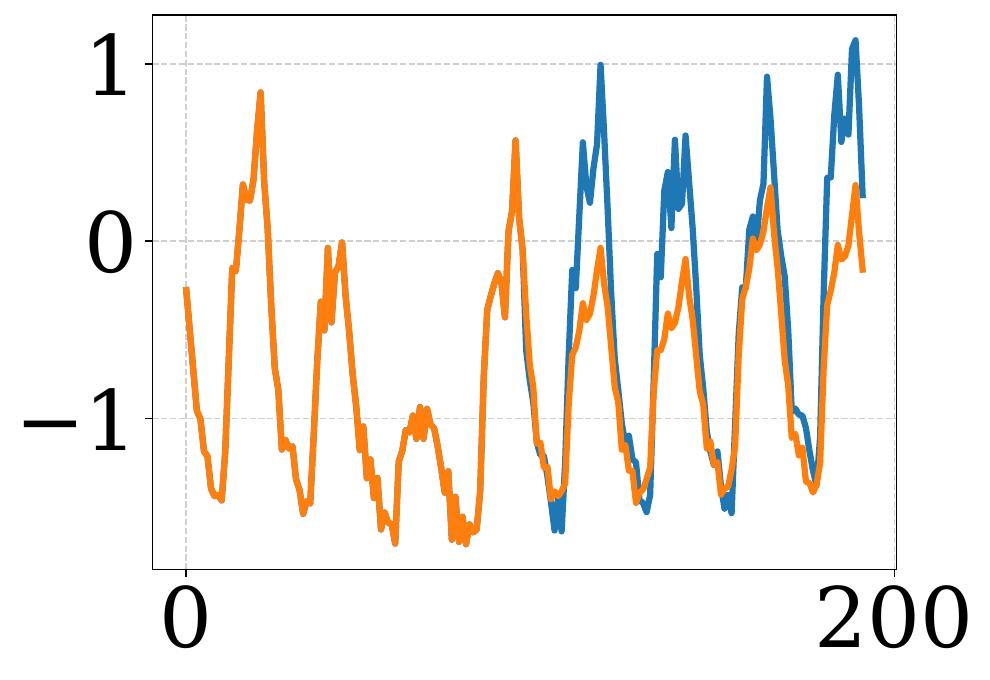} \\
        (a) iTransformer & (b) TimeXer & (c) TimeMixer \\
        \includegraphics[width=0.28\columnwidth]{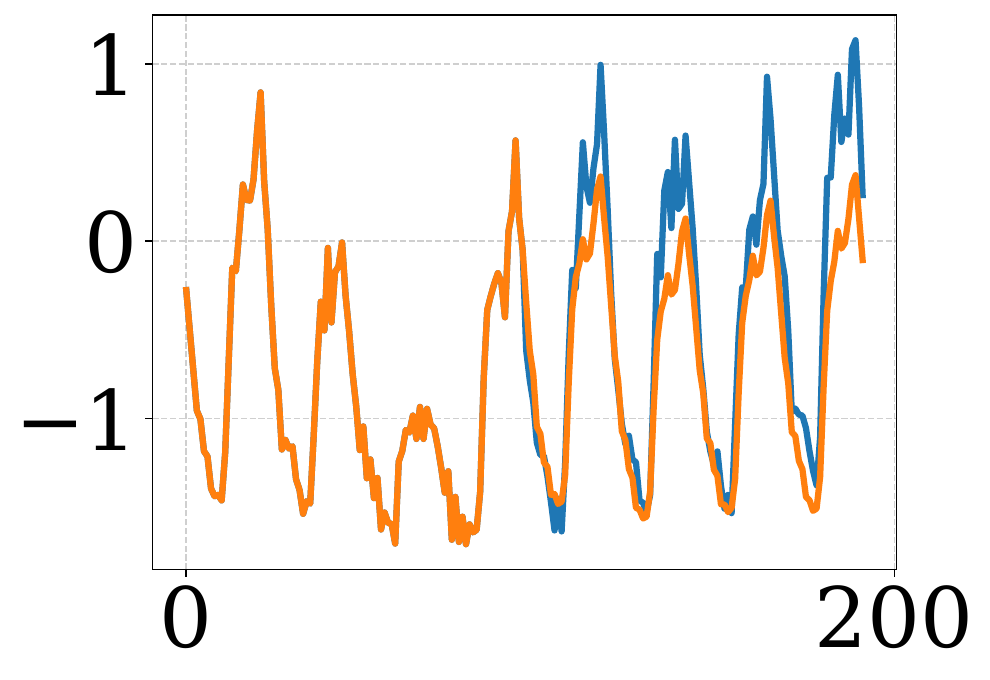} &
        \includegraphics[width=0.28\columnwidth]{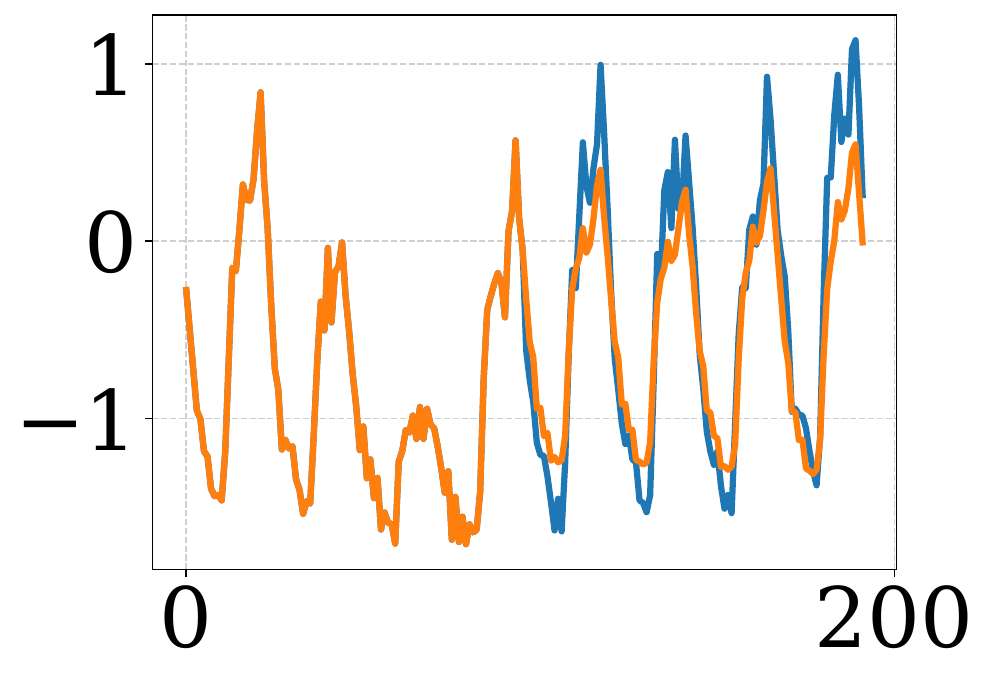} &
        \includegraphics[width=0.28\columnwidth]{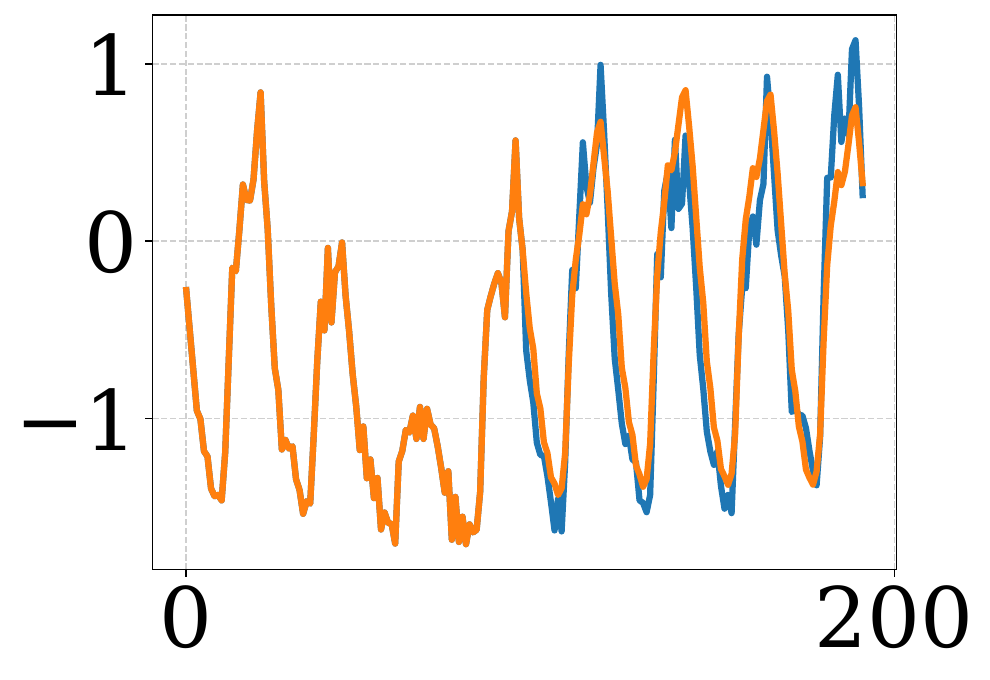} \\
        (d) PatchTST & (e) DLinear & (f) XLinear \\
    \end{tabular}
    \begin{picture}(180,22)(0,0) 
        \textcolor{gray!30}{%
            \put(0,2){\line(1,0){180}}   
            \put(0,20){\line(1,0){180}}  
            \put(0,2){\line(0,1){18}}    
            \put(180,2){\line(0,1){18}}  
        }
        
        \def\legendLineWidth{1.2}
        \def\legendFont{\footnotesize}

        \textcolor[RGB]{30,118,179}{%
            \put(6,12){\line(1,0){25}} 
        }
        \put(36,8){\legendFont GroundTruth} 
        
        \textcolor[RGB]{255,126,13}{%
            \put(95,12){\line(1,0){25}}
        }
        \put(125,8){\legendFont Prediction}
    \end{picture}
    \caption{Typical forecast results of XLinear and 5 SOTA baselines on the last variable of the Electricity dataset (Input and prediction horizons are 96).}
    \label{fig:data_ecl}
\end{figure}
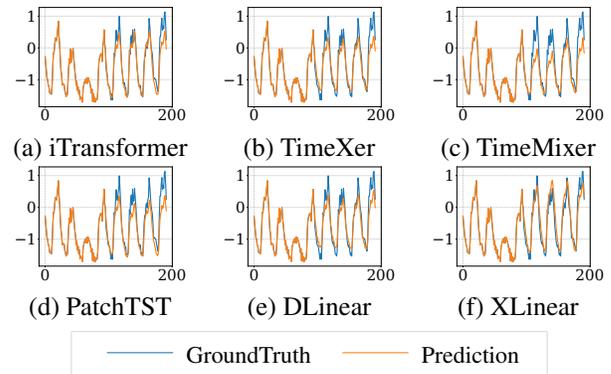

\noindent\textbf{Long-term Multivariate Forecasting.} The forecasting results for MSE and MAE across 7 benchmark datasets are presented in Table \ref{tab:multi_forcasting}. On the first six datasets, XLinear achieves the best performance in over 91.7\% of the cases for both metrics, and ranks second in the remaining cases.

The only exception is on the Traffic dataset with 862 variables, where iTransformer outperforms XLinear due to its variable-wise attention mechanism, which is more effective for high-dimensional forecasting tasks. In this scenario, each variable simultaneously serves as a prediction target and an exogenous driver, leading to a substantial expansion of the VGM input dimensionality. Extracting informative signals while suppressing noise in such high-dimensional spaces poses a significant challenge. Future work will focus on reducing the VGM input dimensionality while preserving essential inter-variable dependencies, thereby enhancing the model’s scalability and performance on high-dimensional datasets. Averaging over the four forecast horizons, XLinear has the lowest MSE and MAE on the first six data sets, and ranks second in MSE and third in MAE on Traffic. Across all 7 benchmarks, XLinear consistently outperforms the four GNN-based models (in the appendix).

\begin{figure}[t]
    \centering
    \includegraphics[width=0.47\textwidth]{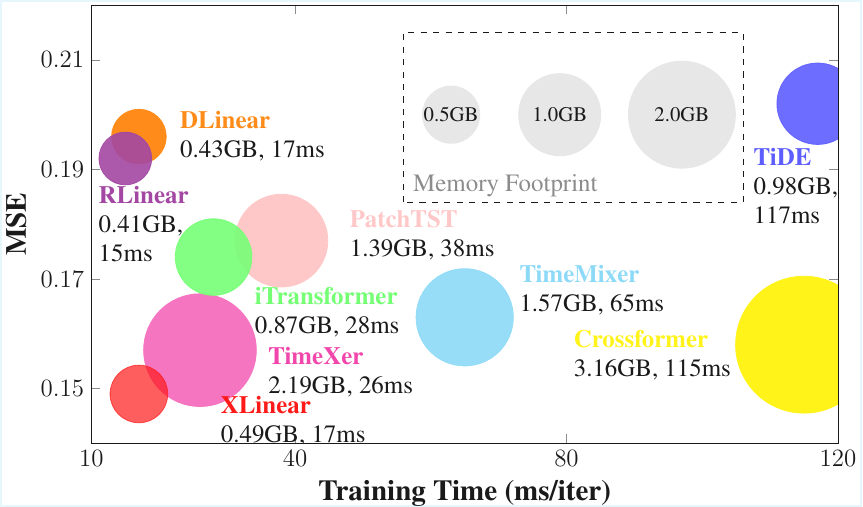}
    \caption{Model analysis of multivariate prediction on the Weather dataset.}
    \label{fig:weather_effience}
\end{figure}

Furthermore, we validate the effectiveness of the XLinear using multivariate forecasting, rather than univariate, to more clearly visualise forecast differences. As illustrated in Fig.~\ref{fig:data_ecl}, the model's predictions closely align with the ground truth, confirming its ability to effectively integrate multivariate information for accurate forecasting. We also assessed XLinear in a multivariate prediction setting using the Weather dataset. The experiment adopted a consistent training batch size of 128, with both the input and output windows set to 96, while the remaining parameters remained at their default values. As shown in Fig~\ref{fig:weather_effience}, XLinear demonstrates strong training efficiency and predictive accuracy. Except  DLinear and RLinear, whose MSEs are 31.5\% and 28.9\% higher respectively , XLinear achieves at least 39.3\% faster training, 43.7\% lower memory usage, and 5.1\% lower MSE than the other time series models in the figure. 

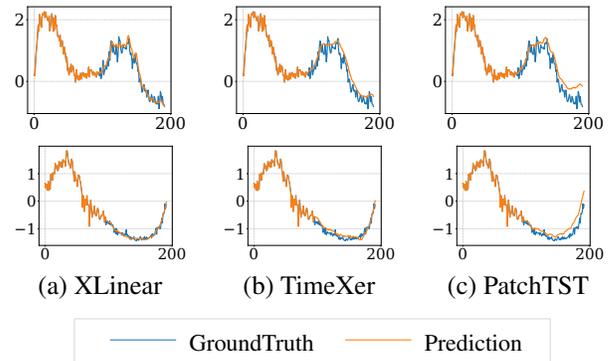
\begin{figure}
    \centering
    \begin{tabular}{ccc}
        \includegraphics[width=0.28\columnwidth]{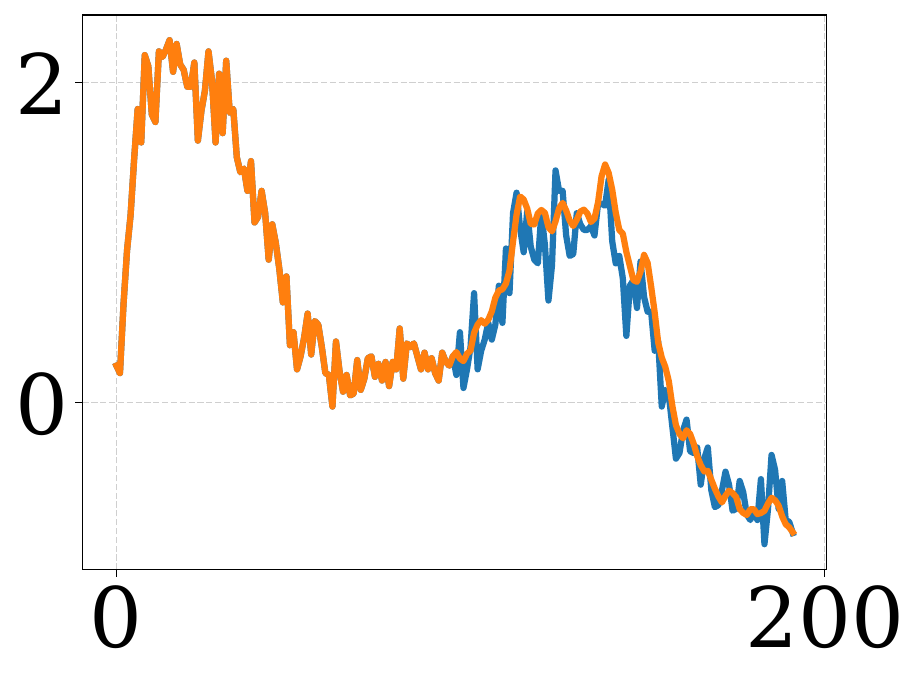} &
        \includegraphics[width=0.28\columnwidth]{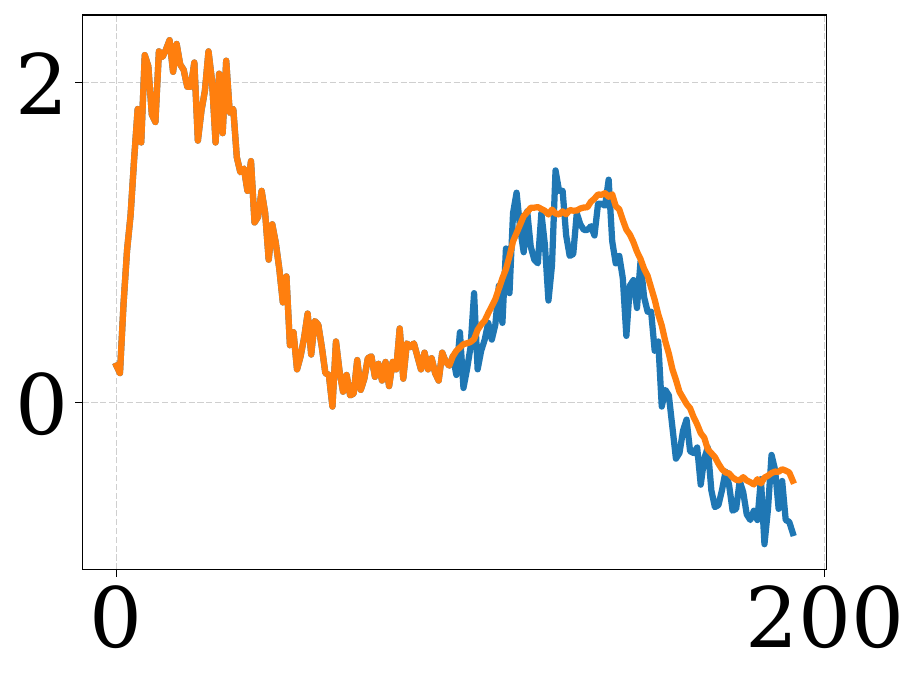} &
        \includegraphics[width=0.28\columnwidth]{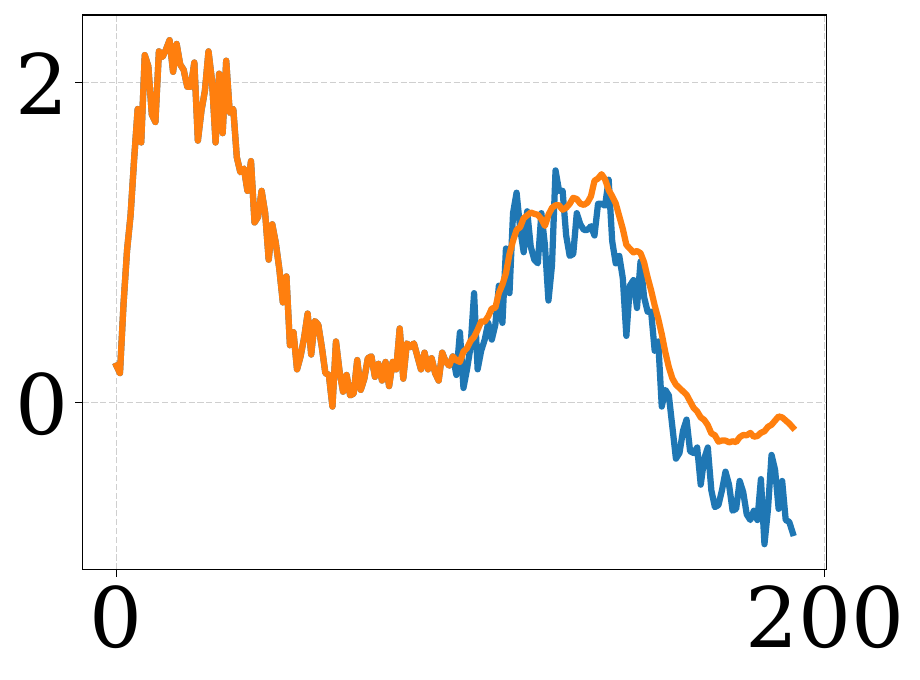} \\
        
        \includegraphics[width=0.28\columnwidth]{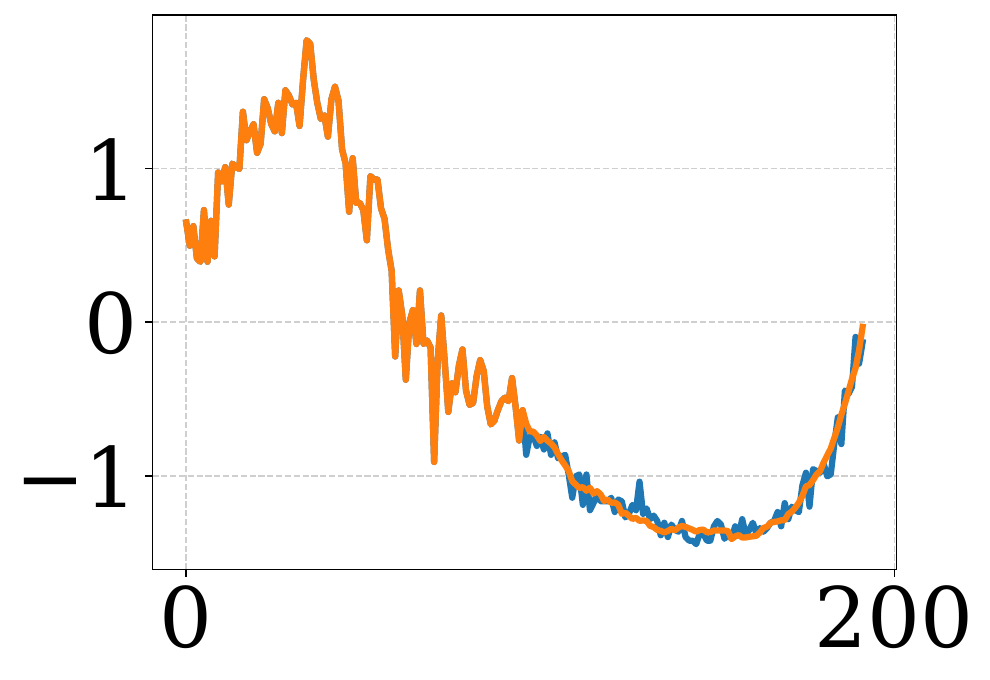} &
        \includegraphics[width=0.28\columnwidth]{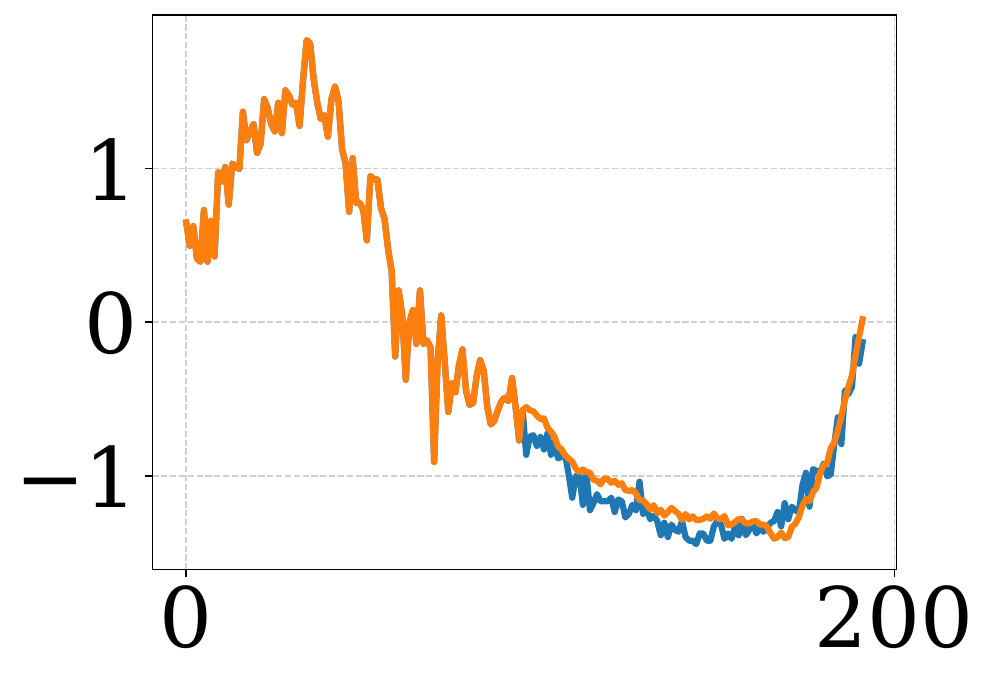} &
        \includegraphics[width=0.28\columnwidth]{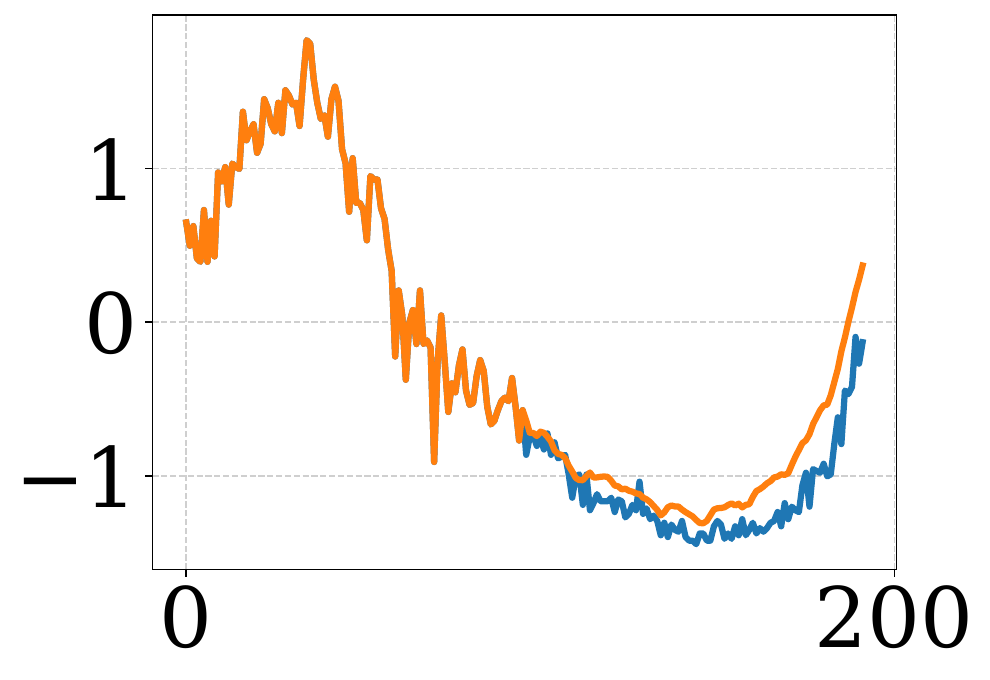} \\
        (a) XLinear & (b) TimeXer & (c) PatchTST \\
    \end{tabular}
    \vspace{3pt} 
    
    \begin{picture}(180,22)(0,0) 
        \textcolor{gray!30}{%
            \put(0,2){\line(1,0){180}}   
            \put(0,20){\line(1,0){180}}  
            \put(0,2){\line(0,1){18}}    
            \put(180,2){\line(0,1){18}}  
        }
        
        \def\legendLineWidth{1.2}
        \def\legendFont{\footnotesize}

        \textcolor[RGB]{30,118,179}{%
            \put(6,12){\line(1,0){25}} 
        }
        \put(36,8){\legendFont GroundTruth} 
        
        \textcolor[RGB]{255,126,13}{%
            \put(95,12){\line(1,0){25}}
        }
        \put(125,8){\legendFont Prediction}
    \end{picture}
    
    \caption{Comparison of prediction performance of various models on PEMS03 (Input and prediction horizons are 96).}
    \label{fig:pems03}
\end{figure}

\begin{table}[t]
    \small
    \centering
    \begin{tabular}{c|c|cc|cc|cc}
        \toprule
        \multicolumn{2}{c|}{Models} & \multicolumn{2}{c|}{{XLinear}} & \multicolumn{2}{c|}{TimeXer} & \multicolumn{2}{c}{PacthTST} \\
        \midrule
        \multicolumn{2}{c|}{Metrics} & MSE & MAE & MSE & MAE & MSE & MAE\\
        \midrule
        \multirow{4}{*}{\rotatebox[origin=c]{90}{PEMS03}} & 12 & \textbf{0.030} & \textbf{0.128} & \underline{0.033} & \underline{0.136} & 0.036 & 0.139 \\
        & 24 & \textbf{0.044} & \textbf{0.152} & \underline{0.050} & \underline{0.168} & 0.061 & 0.170\\
        & 48 & \textbf{0.075} & \textbf{0.190} & \underline{0.086} & \underline{0.217} & 0.125 & 0.233\\
        & 96 & \underline{0.133} & \textbf{0.243} & \textbf{0.129} & \underline{0.264} & 0.226 & 0.321\\
        \midrule
        \multirow{4}{*}{\rotatebox[origin=c]{90}{PEMS04}} & 12 & \textbf{0.047} & \textbf{0.163} & \underline{0.051} & \underline{0.170} & 0.055 & 0.173 \\
        & 24 & \textbf{0.056} & \textbf{0.178} & \underline{0.064} & \underline{0.192} & 0.069 & 0.194\\
        & 48 & \textbf{0.072} & \textbf{0.198} & \underline{0.086} & \underline{0.219} & 0.105 & 0.241\\
        & 96 & \textbf{0.108} & \textbf{0.246} & \underline{0.134} & \underline{0.272} & 0.180 & 0.309\\
        
        \midrule
        \multirow{4}{*}{\rotatebox[origin=c]{90}{PEMS07}} & 12 & \textbf{0.075} & \textbf{0.200} & \underline{0.078} & \underline{0.208} & 0.093 & 0.220 \\
        & 24 & \textbf{0.094} & \textbf{0.225} & \underline{0.096} & \underline{0.234} & 0.133 & 0.264\\
        & 48 & \underline{0.128} & \underline{0.266} & \textbf{0.124} & \textbf{0.264} & 0.208 & 0.332\\
        & 96 & \textbf{0.146} & \textbf{0.289} & \underline{0.158} & \underline{0.304} & 0.287 & 0.380\\
        \midrule
        \multirow{4}{*}{\rotatebox[origin=c]{90}{PEMS08}} & 12 & \textbf{0.153} & \textbf{0.277} & \underline{0.158} & \underline{0.281} & 0.195 & 0.307 \\
        & 24 & \textbf{0.189} & \textbf{0.305} & \underline{0.200} & \underline{0.317} & 0.262 & 0.357\\
        & 48 & \textbf{0.238} & \textbf{0.350} & \underline{0.258} & \underline{0.361} & 0.378 & 0.442\\
        & 96 & \textbf{0.287} & \textbf{0.402} & \underline{0.322} & \underline{0.425} & 0.490 & 0.580\\
        \bottomrule
        
    \end{tabular}
    \caption{Main results of the PEMS forecasting task. }
    \label{lab:pems}
\end{table}

\subsection{Model Analysis}

\noindent\textbf{Comparison with Patch-based Models.}
To mitigate the inherent limitations of the attention mechanism~\cite{zeng2023transformers}, many efficient time series forecasting models have adopted patch-based attention mechanisms. While this approach effectively preserves local semantic information, it reduces sensitivity to fine-grained positional cues, making it challenging to model rapid fluctuations. To test this hypothesis, we conducted comparative experiments with TimeXer and PatchTST on the highly volatile PEMS datasets, under the setting of single endogenous variable forecasting with exogenous variables. Two representative examples are visualized in Fig. \ref{fig:pems03}, and the detailed results are provided in Table \ref{lab:pems}, where XLinear demonstrates superior accuracy across most scenarios. All models can capture the overall trend reasonably well. However, in terms short-term fluctuations, XLinear more precisely forecasts fine-grained changes in the sequences. This advantage stems from its direct modeling of individual dimensions after data embedding, whereas the patching operations in patch-based models may hinder the learning of high-resolution temporal details.

\begin{figure}[t]
    \centering
    \includegraphics[width=0.48\textwidth]{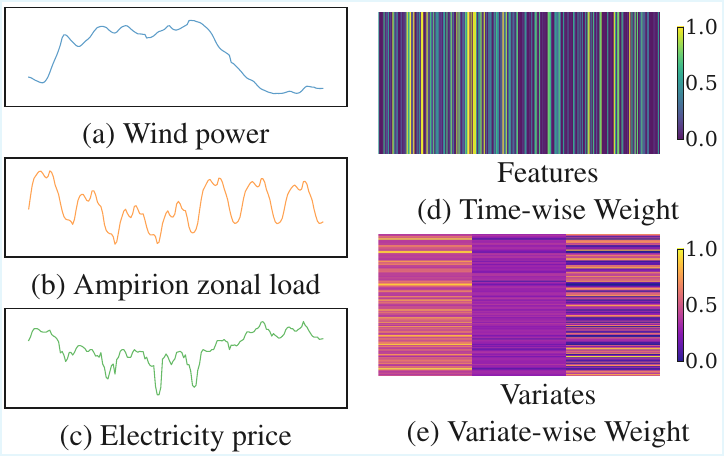}
    \caption{Visualization results of input samples and weights of XLinear on the DE dataset.}
    \label{fig:weight_visual}
\end{figure}

\noindent\textbf{Weight Visualization.}
To investigate the interpretability of XLinear, we selected a representative scenario dataset DE (a subset of the EPF dataset) \cite{lago2021forecasting} and used exogenous variables such as Wind power and Ampirion zonal load to predict Electricity price. Figure \ref{fig:weight_visual} presents a sample cases and their corresponding weights: (a)-(c) the original sequences, which show a significant correlation between Wind power and Electricity price across multiple time intervals; (e) the variable-wise weight distribution, where Channel 1 (representing Wind power) and Channel 3 (representing Electricity price) exhibit higher weights, consistent with the correlations observed in the input sequences; (d) Temporal weight distribution: In this process, XLinear identifies and enhances important features in the original sequences while establishing mapping relationships with global tokens.
Overall, these observations demonstrate that XLinear can effectively highlight key variables and critical time steps, providing clear interpretability of its predictions.

\begin{figure}[t]
    \centering
    \includegraphics[width=0.38\textwidth]{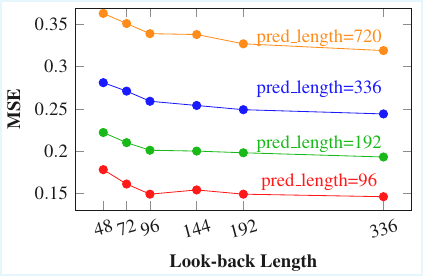}
    \caption{Visualization results of input samples and weights of XLinear on the DE dataset.}
    \label{fig:input_var}
\end{figure}

\noindent\textbf{Long lookback window.} 
Our proposed MLP-based model, XLinear, with its ``time-step-dependent" captured weights, can efficiently extract valid information from longer sequences without temporal attention. As shown in Fig. \ref{fig:input_var},  its forecast MSEs for different prediction lengths drop rapidly as the lookback length increases to 96, and then decreases gradually with further increases. This indicates that XLinear, like most models, benefits from longer input sequences.

\begin{figure}[t]
    \centering
    \includegraphics[width=0.45\textwidth]{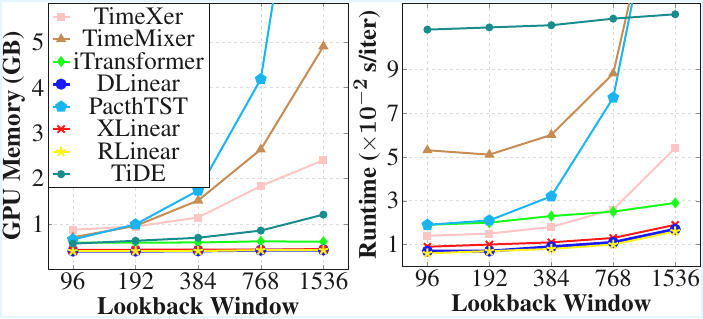}
    \caption{Model efficiency vs. lookback window for multivariate forecasting on the Weather dataset.}
    \label{fig:input-time-memory}
\end{figure}

Furthermore, we investigate how model efficiency varies with the lookback window. On the Weather dataset, the experimental results are shown in Fig. \ref{fig:input-time-memory}, with the training batch size fixed at 32 and the forecast horizon set to 96. These demonstrate that XLinear outperforms all baseline models in terms of training time and memory usage, except being comparable to DLinear and RLinear. Notably, its training time and GPU memory consumption grow slowly and approximately linearly with the lookback window, whereas the resource demands of TimeMixer and PatchTST increase at least quadratically. This clearly demonstrates the high computational efficiency of XLinear when handling long input sequences. 

To assess the effectiveness of XLinear in long-input scenarios, we conducted more comparative experiments with state-of-the-art models such as TSMixer and PatchTST, using a lookback window of length 512 for multivariate forecasting tasks. As presented in Table \ref{lab:compare_with_tsmixer}, for forecast horizons ranging from 96 to 336 steps, XLinear effectively leverages the long lookback window and consistently outperforms both TSMixer and PatchTST. At the longest horizon of 720 steps, XLinear performs comparably to PatchTST, which begins to show a general advantage over TSMixer. These results indicate that XLinear is better able to utilise long lookback windows than TSMixer. 

\begin{table}[t]
    \centering
    \small
    \begin{tabular}{c|c|cc|cc|cc}
        \toprule
        \multicolumn{2}{c|}{Models} & \multicolumn{2}{c|}{{XLinear}} & \multicolumn{2}{c|}{TSMixer} & \multicolumn{2}{c}{PacthTST} \\
        \midrule
        \multicolumn{2}{c|}{Metrics} & MSE & MAE & MSE & MAE & MSE & MAE\\
        \midrule
        \multirow{4}{*}{\rotatebox[origin=c]{90}{ETTh1}}
        & 96 & \textbf{0.357} & \textbf{0.389} & \underline{0.361} & \underline{0.392} & 0.370 & 0.400 \\
        & 192 & \textbf{0.393} & \textbf{0.411} & \underline{0.404} & \underline{0.418} & 0.413 & 0.429\\
        & 336 & \textbf{0.397} & \textbf{0.425} & \underline{0.420} & \underline{0.431} & 0.422 & 0.440\\
        & 720 & \underline{0.451} & \underline{0.470} & 0.463 & 0.472 & \textbf{0.447} & \textbf{0.468}\\
        \midrule
        \multirow{4}{*}{\rotatebox[origin=c]{90}{ETTh2}}
        & 96 & \textbf{0.258} & \textbf{0.328} & \underline{0.274} & 0.341 & \underline{0.274} & \underline{0.337} \\
        & 192 & \textbf{0.314} & \textbf{0.367} & \underline{0.339} & 0.385 & 0.341 & \underline{0.382}\\
        & 336 & \textbf{0.320} & \textbf{0.382} & 0.361 & 0.406 & \underline{0.329} & \underline{0.384}\\
        & 720 & \underline{0.385} & \underline{0.427} & 0.445 & 0.470 & \textbf{0.379} & \textbf{0.422}\\
        \midrule
        \multirow{4}{*}{\rotatebox[origin=c]{90}{ETTm1}}
        & 96 & \textbf{0.280} & \textbf{0.336} & \underline{0.285} & \underline{0.339} & 0.293 & 0.346 \\
        & 192 & \textbf{0.325} & \textbf{0.364} & \underline{0.327} & \underline{0.365} & 0.333 & 0.370\\
        & 336 & \textbf{0.354} & \textbf{0.382} & \underline{0.356} & \underline{0.382} & 0.369 & 0.392\\
        & 720 & \textbf{0.415} & \textbf{0.414} & 0.419 & \textbf{0.414} & \underline{0.416} & \underline{0.420}\\
        \midrule
        \multirow{4}{*}{\rotatebox[origin=c]{90}{Weather}}
        & 96 & \textbf{0.142} & \textbf{0.195} & \underline{0.145} & \underline{0.198} & 0.149 & \underline{0.198} \\
        & 192 & \textbf{0.185} & \textbf{0.235} & \underline{0.191} & 0.242 & 0.194 & \underline{0.241}\\
        & 336 & \textbf{0.237} & \textbf{0.276} & \underline{0.242} & \underline{0.280} & 0.245 & 0.282\\
        & 720 & \textbf{0.311} & \textbf{0.330} & 0.320 & 0.336 & \underline{0.314} & \underline{0.334}\\
        \midrule
    \end{tabular}
    \caption{Performance comparison with a lookback window length of 512.}
    \label{lab:compare_with_tsmixer}
\end{table}

\subsection{Ablation Study}  
\begin{table}[t]
    \centering
    \small
    \begin{tabular}{c|c|cc|cc|cc}
        \toprule
        \multicolumn{2}{c|}{Models} & \multicolumn{2}{c|}{{XLinear}} & \multicolumn{2}{c|}{ES} & \multicolumn{2}{c}{GT} \\
        \midrule
        \multicolumn{2}{c|}{Metrics} & MSE & MAE & MSE & MAE & MSE & MAE\\
        \midrule
        \multirow{4}{*}{\rotatebox[origin=c]{90}{ETTh1}} & 96 & \textbf{0.369} & \textbf{0.393} & 0.370 & 0.396 & 0.382 & 0.406 \\
        & 192 & \textbf{0.421} & \textbf{0.423} & 0.427 & 0.427 & 0.429 & 0.430\\
        & 336 & \textbf{0.455} & \textbf{0.440} & 0.462 & 0.444 & 0.467 & 0.449\\
        & 720 & \textbf{0.453} & \textbf{0.456} & 0.461 & 0.461 & 0.475 & 0.471\\
        \midrule
        \multirow{4}{*}{\rotatebox[origin=c]{90}{ETTm1}} & 96 & \textbf{0.311} & \textbf{0.351} & 0.328 & 0.366 & 0.326 & 0.364 \\
        & 192 & \textbf{0.353} & \textbf{0.376} & 0.364 & 0.383 & 0.362 & 0.382\\
        & 336 & \textbf{0.382   } & \textbf{0.398} & 0.392 & 0.404 & 0.390 & 0.404\\
        & 720 & \textbf{0.444} & \textbf{0.436} & 0.457 & 0.441 & 0.463 & 0.444\\
        \midrule
        \multirow{4}{*}{\rotatebox[origin=c]{90}{Weather}} & 96 & \textbf{0.149} & \textbf{0.198} & 0.175 & 0.216 & 0.153 & 0.200 \\
        & 192 & \textbf{0.201} & \textbf{0.244} & 0.226 & 0.258 & 0.204 & 0.247\\
        & 336 & \textbf{0.259} & \textbf{0.288} & 0.282 & 0.298 & 0.263 & 0.291\\
        & 720 & \textbf{0.339} & \textbf{0.340} & 0.357 & 0.347 & 0.344 & 0.344\\
        \midrule
        \multirow{4}{*}{\rotatebox[origin=c]{90}{\scalebox{0.9}{GTD$_S$\_{MS}}}} & 96 & \textbf{0.012} & \textbf{0.077} & 0.014 & 0.083 & \textbf{0.012} & 0.078 \\
        & 192 & \textbf{0.025} & \textbf{0.112} & 0.027 & 0.118 & \textbf{0.025} & 0.114\\
        & 336 & \textbf{0.048} & \textbf{0.159} & 0.051 & 0.163 & 0.049 & 0.161\\
        & 720 & \textbf{0.124} & \textbf{0.260} & 0.130 & 0.267 & 0.127 & 0.266\\
        \bottomrule 
    \end{tabular}
    \caption{Results of the ablation experiments, where the optimal results are highlighted in \textbf{bold}.}
    \label{lab:ablation}
\end{table}

In the prediction phase, XLinear generates forecasts by synergistically integrating the temporal dependencies of the endogenous sequence and the key exogenous inputs embedded in the global tokens. To assess the contributions of the two information sources, we conducted ablation experiments under three configurations: (a) using both the endogenous sequence to capture temporal dependencies and the global tokens representing interaction from exogenous variables  (XLinear); (b) using only the endogenous sequence (denoted as ES); (c) using only the global tokens (denoted as GT). As shown in Table \ref{lab:ablation}, integrating both information sources yields the best performance. Notably, the endogenous global token alone generally outperforms the pure temporal sequence, as it captures both the target’s overall temporal features and exogenous information, highlighting their complementary roles.

\section{Conclusion}
Motivated by real-world time series forecasting applications, particularly those involving exogenous inputs, we designed XLinear. This lightweight model incorporates  a gating mechanism, utilizing MLP with sigmoid activation functions. XLinear effectively captures both temporal patterns and variate-wise dependencies from exogenous variables to endogenous variables through their learnable global tokens via time-wise and variate-wise gating modules. These gated outputs are fused with the input feature maps to amplify salient signals and suppress noise. Extensive experiments on seven public benchmarks and five real-world application datasets demonstrated that XLinear achieved superior accuracy and efficiency for both multivariate forecasting and univariate forecasting influenced by exogenous variables. Notably, it matched the speed of other lightweight models while being significantly (at least 30\%) faster than the most efficient Transformer-based alternatives. \\
For future work, we aim to enhance XLinear's scalability to handle time series with hundreds of variables. Promising directions include further improving forecast accuracy by optimizing lookback windows and bolstering forecast reliability for operational deployment.

\section{Acknowledgments}
This research project was supported in part by National Key Research and Development Program of China under Grant 2023YFF1000100, and Hubei Key Research and Development Program of China under Grant 2024BBB055, 2024BAA008; and in part by the Major Science and Technology Project of Yunnan Province under Grant 202502AE090003, and in part by the Fundamental Research Funds for the Chinese Central Universities under Grant 2662025XXPY005. This work was partially supported by CSIRO Digital Water and Landscapes project.

\bibliography{aaai2026}

\section{Appendix}
In the supplementary materials, we will summarize research progress related to our study and present additional experimental results. The structure of the supplementary materials is as follows
\begin{itemize}
\item In Subsection A, we will brief the Related Work.
\item In Subsection B, we describe datasets used.
\item In Subsection C,  we compare the forecast accuracy of XLinear with that of advanced time series models, and GNN-based models.
\item In Subsection D, we report a targeted ablation experiment examining the role of the activation function in the gating module.
\end{itemize}

\subsection{A. Related Work}
Since the advent of the Transformer architecture, numerous studies have focused on optimizing the self-attention mechanism for time series modelling. LogTrans~\cite{li2019enhancing} employs a convolution-based self-attention layer with LogSparse design to capture local information, thereby reducing computational complexity; Informer~\cite{zhou2021informer} proposes a ProbSparse self-attention mechanism integrated with distillation technology, enabling efficient extraction of key key-value pairs; Autoformer~\cite{wu2021autoformer} draws on decomposition and autocorrelation ideas from traditional time series analysis to capture complex temporal patterns. However, the point-wise attention adopted by these models lacks local semantic information, and the permutation invariance of self-attention can result in the loss of temporal information in sequences, allowing even simple linear models~\cite{zeng2023transformers} to potentially outperform these sophisticated Transformer models. To address these limitations, PatchTST~\cite{nie2022time} introduces a patch-based processing approach, which not only preserves local semantic information but also significantly improves computational efficiency; Crossformer~\cite{zhang2023crossformer} further enables interactions across both temporal and variable dimensions to capture more effective information; iTransformer~\cite{liu2023itransformer} alters the operation of the self-attention mechanism through variable-level embeddings to capture inter-variable dependencies; TimeXer~\cite{wang2024timexer} leverages global tokens learned from the temporal dimension and captures effective variate-wise dependencies via cross-attention mechanisms, achieving SOTA forecast accuracy. Patch-based Transformer variants excessively rely on the patching mechanism to achieve desirable performance, which limits their applicability in forecasting tasks unsuitable for patching~\cite{luo2024deformabletst}. Additionally, their permutation-invariant self-attention mechanism may suffer from temporal information loss~\cite{tang2025unlocking}.

MLP-based time series forecasting models are gaining increasing attention due to their excellent balance between predictive accuracy and computational efficiency. These models use lightweight architectures to achieve strong predictive capabilities through efficient feature extraction. DLinear~\cite{zeng2023transformers} decomposes inputs into trend and seasonal components, and then applies separate linear layers to each for efficient long-term time series forecasts. RLinear~\cite{li2023revisiting} further demonstrates that a simple linear mappling, combined with RevIN (Reversible Instance Normalization), can achieve strong performance. TSMixer~\cite{ekambaram2023tsmixer} enhances MLP with gated attention to learn key features within and across patches, capturingcross-variable dependencies through cross-channel coordination heads. TiDE~\cite{das2023long} introduces a dense encoder-decoder based on MLPs for extracting salient temporal features.  TimeMixer~\cite{wang2024timemixer} draws on the season-trend decomposition strategies from Autoformer and DLinear to capture complex temporal patterns at multiple scales. Most recently, xPatch~\cite{stitsyuk2025xpatch} applies exponential moving average to decompose time series into trend and seasonal components, which are then processed independently by two parallel streams: an MLP-based linear stream and a CNN-based non-linear stream. xPatch treats each variable independently and therefore cannot incorporate exogenous inputs. 

In practical forecasting tasks, various scenarios involve multivariate settings that include exogenous variables, such as irrigation scheduling~\citep{shao2019new,shao2025comprehensive}, population flows~\citep{bakar2018spatio}, yield forecast~\citep{jin2022improving}, and environmental science~\citep{genova2025advancing,bakar2015spatiodynamic,kokic2013improved,jin2011towards}. Such tasks require models to capture temporal features in the historical sequences of endogenous variables while fully effectively incorporating exogenous information to assist in forecasting future values of the endogenous variables. Research in this area has gradually advanced. Specifically, TSMixer~\cite{chen2023tsmixer}, designed to fuse heterogeneous features, first projects  features of different types into a unified shape to enable concatenation. During the mixing stage, its mixing layer, which incorporates time-mixing and feature-mixing operations, jointly extracts temporal patterns and cross-variate dependencies, ultimately generating outputs for each time step through a fully connected layer. TiDE~\cite{das2023long} inputs the historical values of both endogenous and exogenous variables into the encoder to extract shared representations, and incorpoerates future values of exogenous variables during decoding to provide auxiliary information for predicting future endogenous values. TimeXer~\cite{wang2024timemixer} first captures temporal dependencies among endogenous variables via self-attention mechanisms, while learning a global token that facilitates interaction with exogenous variables via cross-attention. This mechanism preserves information across both temporal and variable dimensions, thereby enhancing predictive performance. Building on this line of research, our work incorporates MLP-based gating modules to further improve both forecasting efficiency and accuracy.

Pre-trained foundation models for time series forecasting have seen remarkable progress in recent years, with most built upon Transformer architectures~\cite{kottapalli2025foundation}.However, their forecasting accuracy often falls short on new or unseen datasets, and even on some in-domain tasks~\cite{shi2025time}, when compared to full-shot state-of-the-art models such as TimeXer~\cite{wang2024timemixer} and TimeMixer~\cite{wang2024timemixer}. In addition, other deep learning approaches have been explored, including convolution-based models~\cite{wu2022timesnet} and graph neural networks (GNNs)~\cite{huang2023crossgnn,cai2024msgnet,wu2020connecting,yi2023fouriergnn}. These models face two major limitations: they typically lack explicit mechanisms to incorporate exogenous variables, and their forecasting performance on multivariate time series remains inferior to both TimeXer~\cite{wang2024timexer} and our proposed model, XLinear.

\subsection{B. Datasets}
In this paper, we primarily incorporate the following 12 datasets to evaluate the performance of XLinear:
\begin{itemize}
    \item Electricity Transformer Temperature (ETT)\footnote{https://github.com/zhouhaoyi/ETDataset}: comprises 4 subsets, with data sourced from two distinct regions in China over a 2-year time span. Among them, {ETTh1, ETTh2} are datasets with a 1-hour sampling granularity, while {ETTm1, ETTm2} are datasets with a 15-minute sampling granularity. Each data sample contains the target variable ``oil temperature" and 6 (exogenous) variables related to power load.
    \item Weather\footnote{https://www.bgc-jena.mpg.de/wetter}: collects 21 meteorological indicators in Germany, such as humidity and air temperature, collected every 10 minutes from the Weather Station of the Max Planck Biogeochemistry Institute in 2020.
    \item Electricity\footnote{https://archive.ics.uci.edu/dataset/321/}: includes the hourly electricity consumption data of 321 customers spanning from 2012 to 2014.
    \item Traffic\footnote{https://pems.dot.ca.gov}: consists of hourly data collected by the California Department of Transportation, which records the road occupancy rates measured by various sensors on freeways in the San Francisco Bay Area.
    \item Dissolved Oxygen\footnote{realtimedata.waternsw.com.au/water.stm}\textsuperscript{,}\footnote{data.water.vic.gov.au/WMIS/}: Contains dissolved oxygen data from two monitoring stations (with identifiers DO$_{425012}$ and DO$_{409215}$ respectively), corresponding to the Darling River (upstream of Weir 32) and the Murray River (Tocumwal) in Australia. This dataset records environmental factors (such as water temperature, mean discharge rate) and water quality indicators including dissolved oxygen every 15 minutes.
    \item Water temperature: collected from the northern and southern monitoring sites of Grahamstown Dam, Australia (denoted as GTD$_N$ and GTD$_S$ respectively), it records environmental factors and water temperatures at 0.5-9 meters underwater at 1-hour intervals. Access to these two datasets may be granted upon request.
    \item Crop Yield Prediction\footnote{https://www.kaggle.com/datasets/ajithdari/crop-yield-prediction}: contains environmental data collected from agricultural fields at 5-minute intervals, including soil humidity, air temperature, air humidity, wind speed, and other weather parameters. It also inlcudes crop yield measurements collected between February 23 and March 25, 2019.
\end{itemize}

\begin{table*}[htbp]
    \centering
    \scriptsize
    \begin{tabular}{ccccccccccccccc} 
        \toprule
         Model & \multicolumn{2}{c}{ECL} & \multicolumn{2}{c}{Weather} & \multicolumn{2}{c}{ETTh1} & \multicolumn{2}{c}{ETTh2} & \multicolumn{2}{c}{ETTm1} & \multicolumn{2}{c}{ETTm2} & \multicolumn{2}{c}{Traffic} \\ 
         \cmidrule(lr){2-3}\cmidrule(lr){4-5}\cmidrule(lr){6-7}\cmidrule(lr){8-9}\cmidrule(lr){10-11}\cmidrule(lr){12-13}\cmidrule(lr){14-15}
         Metric & MSE & MAE & MSE & MAE & MSE & MAE & MSE & MAE & MSE & MAE & MSE & MAE & MSE & MAE \\ \midrule
        XLinear & \textbf{0.168} & \textbf{0.263} & \textbf{0.237} & \textbf{0.268} & \textbf{0.425} & \textbf{0.428} & \textbf{0.359} & \textbf{0.391}  & \textbf{0.373} & \textbf{0.390} & \textbf{0.270} & \textbf{0.316}  & \underline{0.463} & {0.295}  \\ \midrule
        TimeXer & \underline{0.171} & \underline{0.270} & 0.241 & \underline{0.271} & \underline{0.437} & \underline{0.437}  & 0.366 & \underline{0.395}  & 0.382 & 0.397 & \underline{0.274} & \underline{0.322} & 0.467 & \underline{0.288} \\ \midrule
        TimeMixer & 0.182 & 0.272 & \underline{0.240} & \underline{0.271} & 0.447 & 0.440  & \underline{0.364} & \underline{0.395}  & \underline{0.381} & \underline{0.395} & 0.275 & 0.323 & 0.484 & 0.297  \\ \midrule
        iTransformer & 0.178 & \underline{0.270} & 0.258 & 0.279 & 0.454 & 0.447  & 0.383 & 0.407  & 0.407 & 0.410 & 0.288 & 0.332 & \textbf{0.428} & \textbf{0.282}  \\ \midrule
        MTGNN \cite{wu2020connecting} & 0.251 & 0.347 & 0.314 & 0.355 & 0.572 & 0.553          & 0.465 & 0.509  & 0.468 & 0.446 & 0.324 & 0.365 & 0.650 &0.446            \\ \midrule
        CrossGNN \cite{huang2023crossgnn}  & 0.201 & 0.271 & 0.247 & 0.289 & 0.437 & 0.434 & 0.363 & 0.418 & 0.393 & 0.404 & 0.282 & 0.330 & 0.583 & 0.323 \\ \midrule
        MSGNet \cite{cai2024msgnet} & 0.194 & 0.300 & 0.249 & 0.278 & 0.0.452 & 0.452 & 0.396 & 0.417 & 0.398 & 0.411 & 0.288 & 0.330 & - & -            \\ \midrule
        FourierGNN \cite{yi2023fouriergnn} & 0.228 & 0.324 & 0.249 & 0.302 & - & - & - & - & - & - & - & - & 0.557 & 0.342            \\
        \bottomrule
        \end{tabular}
    \caption{Comparison with GNN-based models. All baseline results, except for TimeMixer, are reported in~\cite{wang2024timexer}. ‘-’ denotes there is no officially reported results.}
    \label{gnn-based}
\end{table*}
\subsection{C. Comparison with GNN-based Models} As a class of potential solutions, some Graph Neural Network (GNN)-based models were included in the baseline models for comprehensive comparison. As shown in Table \ref{gnn-based}, we conducted comparative experiments on 7 public datasets in multivariate prediction settings. The average MSE results across four forecast horizons indicate that XLinear exhibits significant advantages over these GNN-based models across all datasets 

Considering all the SOTA models. including TimeXer, TimeMixer, and iTransformer, XLinear achieves the lowest MAE and MSE on all dataset except for Traffic. On the high-dimensional Traffic dataset, XLinear ranks second in MSE and third in MAE.

\begin{table*}[t]
    \centering
    \scalebox{0.7}{
        \begin{tabular}{c|c|cc|cc|cc|cc}
            \toprule[2pt]
            \multicolumn{2}{c|}{Activation} & \multicolumn{2}{c|}{{Sigmoid}} & \multicolumn{2}{c|}{Swish} & \multicolumn{2}{c|}{Tanh} & \multicolumn{2}{c}{Softmax}\\
            \midrule
            \multicolumn{2}{c|}{Metrics} & MSE & MAE & MSE & MAE & MSE & MAE & MSE & MAE\\
            \midrule
            \multirow{4}{*}{\rotatebox[origin=c]{90}{ETTh1}}
            & 96 & \textbf{0.369} & \textbf{0.393} & 0.375 & 0.398 & 0.385 & 0.403 & 0.410 & 0.422\\
            & 192 & \textbf{0.421} & \textbf{0.423} & 0.430 & 0.429 & 0.431 & 0.429 & 0.461 & 0.451\\
            & 336 & \textbf{0.455} & \textbf{0.440} & 0.467 & 0.446 & 0.468 & 0.444 & 0.487 & 0.464\\
            & 720 & \textbf{0.453} & \textbf{0.456} & 0.466 & 0.460 & 0.480 & 0.469 & 0.473 & 0.468\\
            \midrule
            \multirow{4}{*}{\rotatebox[origin=c]{90}{ETTh2}}
            & 96 & \textbf{0.286} & \textbf{0.337} & 0.289 & 0.339 & 0.294 & 0.342 & 0.329 & 0.374\\
            & 192 & \textbf{0.363} & \textbf{0.388} & \textbf{0.363} & \textbf{0.388} & 0.372 & 0.393 & 0.412 & 0.421\\
            & 336 & \textbf{0.378} & \textbf{0.407} & 0.386 & 0.412 & 0.391 & 0.416 & 0.410 & 0.431\\
            & 720 & \textbf{0.408} & \textbf{0.431} & 0.424 & 0.440 & 0.432 & 0.441 & 0.434 & 0.449\\
            \midrule
            \multirow{4}{*}{\rotatebox[origin=c]{90}{ETTh1\_{ms}}}
            & 96 & \textbf{0.055} & \textbf{0.178} & 0.056 & 0.180 & 0.056 & 0.180 & 0.060 & 0.186\\
            & 192 & \textbf{0.071} & \textbf{0.202} & \textbf{0.071} & 0.203 & \textbf{0.071} & \textbf{0.202} & 0.080 & 0.218\\
            & 336 & \textbf{0.084} & \textbf{0.226} & 0.085 & 0.229 & \textbf{0.084} & 0.227 & 0.092 & 0.239\\
            & 720 & \textbf{0.083} & \textbf{0.227} & 0.084 & 0.228 & 0.084 & 0.229 & 0.102 & 0.252\\
            \midrule
            \multirow{4}{*}{\rotatebox[origin=c]{90}{ETTh2\_{ms}}}
            & 96 & \textbf{0.130} & \textbf{0.277} & 0.134 & 0.282 & 0.135 & 0.283 & 0.152 & 0.304\\
            & 192 & \textbf{0.180} & \textbf{0.331} & 0.182 & 0.334 & 0.181 & 0.333 & 0.213 & 0.367\\
            & 336 & \textbf{0.209} & \textbf{0.365} & 0.215 & 0.369 & 0.220 & 0.373 & 0.244 & 0.398\\
            & 720 & \textbf{0.217} & \textbf{0.373} & 0.225 & 0.380 & 0.235 & 0.388 & 0.294 & 0.440\\
            \midrule
        \end{tabular}
        }
    \caption{Performance of XLinear with different gating activation functions.}
    \label{lab:ablation study on activation}
\end{table*}
\subsection{D. Ablation Study on Activation Functions}
In the two gating modules, the MLP captures inter-dimensional dependencies within the sequences and generates a weight distribution, which is subsequently mapped to a stable range of (0, 1) via the sigmoid activation. The sigmoid provides smooth and continuous modulation of feature responses, enabling adaptive reweighting and selective scaling based on feature importance. To further investigate the effect of gating activation design, we replace the sigmoid with alternative activations—Swish, Tanh, and Softmax—to evaluate the performance of XLinear.The experimental results are summarized in Table \ref{lab:ablation study on activation}.
During the experiments, all other parameters were kept constant, and replacing the activation function with Swish or Tanh resulted in a slight performance decline in certain cases. In contrast, using Softmax as the activation function led to a substantial deterioration in performance. This is primarily because Softmax imposes a global constraint on the feature space, with its output weights dependent on the entire feature vector, thereby limiting the model's ability to emphasize important features while suppressing noisy ones.

\end{document}